\documentclass[10pt,twocolumn,letterpaper]{article}

\usepackage{cvpr}              

\usepackage{docmute}

\usepackage{graphicx}

\usepackage{xcolor}
\usepackage{colortbl}

\usepackage{multirow}
\usepackage{tabularx}

\usepackage{booktabs}
\usepackage{longtable}
\usepackage{makecell}
\usepackage{caption}

\usepackage{tcolorbox}
\tcbuselibrary{breakable}
\usepackage{algorithm}
\usepackage{algorithmic}
\usepackage{microtype}
\usepackage{titletoc}
\usepackage{soul}

\definecolor{tabcolor3}{RGB}{210, 231, 209} 
\definecolor{tabcolor5}{RGB}{212, 220, 248} 
\definecolor{casebg}{RGB}{255, 250, 240}    

\newcommand\DoToC{%
  \startcontents
  \printcontents{l}{1}{\textbf{Contents}\vskip3pt\hrule\vskip5pt}
  \vskip3pt\hrule\vskip5pt
}

\definecolor{cvprblue}{rgb}{0.21,0.49,0.74}
\usepackage[pagebackref,breaklinks,colorlinks,allcolors=cvprblue]{hyperref}

\def\paperID{4771} 
\def\confName{CVPR}
\def\confYear{2026}

\newcommand{\cebigger}{\includegraphics[width=1.8cm,height=1.8cm]{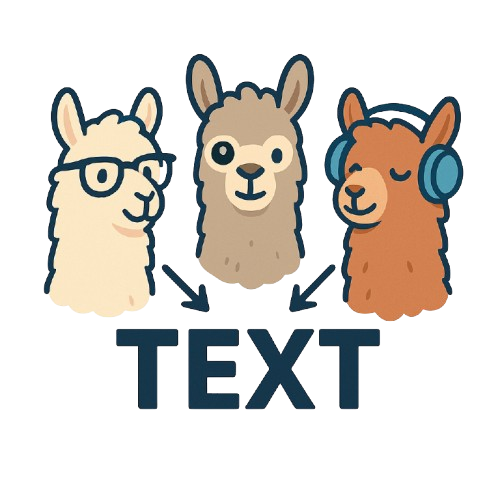}}
\title{
  \begin{tabular}{c}
    \begin{tabular}{@{}l@{\hspace{0.3cm}}c@{}}
      \raisebox{-0.5\height}{\cebigger} & 
      \parbox[t]{0.75\textwidth}{\centering Consensus Entropy: Harnessing Multi-VLM Agreement for Self-Verifying and Self-Improving OCR}
    \end{tabular}
  \end{tabular}
  \vspace{-1.0em}
}

\author{
  Yulong Zhang$^{1,2,3}$\textsuperscript{*},
  Tianyi Liang$^{2,4}$\textsuperscript{*},
  Xinyue Huang$^{5}$,
  Erfei Cui$^{3}$,
  Guoqing Wang$^{4}$,
  Xu Guo$^{1,2}$,\\
  Chenhui Li$^{4}$,
  Gongshen Liu$^{3}$\textsuperscript{†} \\
  $^{1}$Fudan University
  $^{2}$Shanghai Innovation Institute \\
  $^{3}$Shanghai Jiao Tong University
  $^{4}$East China Normal University
  $^{5}$Sun Yat-sen University
  \vspace{-1.5em}
  }

\begin{document}
\twocolumn[{%
\renewcommand\twocolumn[1][]{#1}%
\maketitle
\begin{center}
    \centering
    \includegraphics[width=0.92\textwidth]{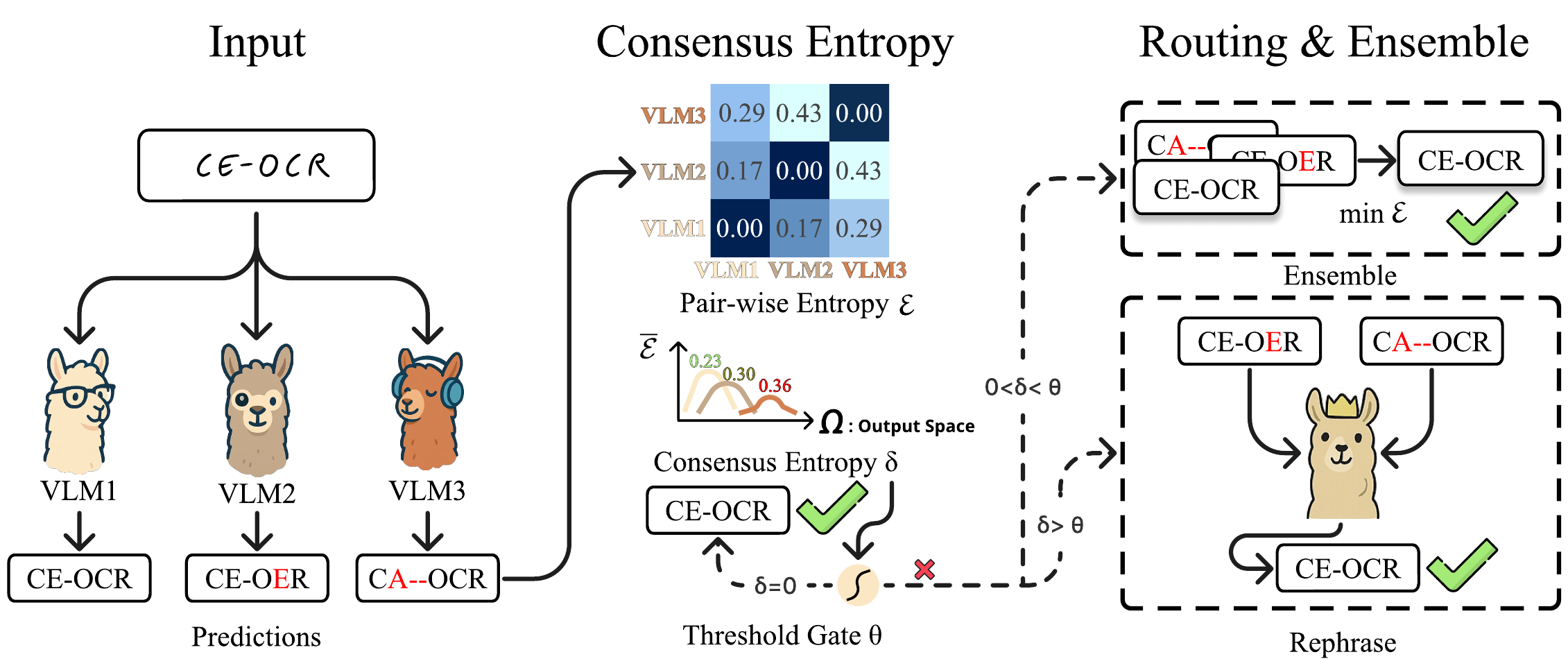}
    \captionsetup{type=figure}
    \captionof{figure}{\textbf{Overview of the CE-OCR Framework.}
    Given an input image, multiple Vision-Language Models (VLMs) independently generate OCR predictions.
    Pairwise similarities among these results yield a probability distribution over consensus quality, from which the \textit{Consensus Entropy} $\delta$ is derived.
    Based on $\delta$, a threshold gate $\theta$ determines the next step:
    low-entropy ensemble predictions are accepted, while inputs with entropy exceeding the threshold are routed to a stronger VLM for rephrasing.
    This framework enables automatic self-verification and quality improvement of OCR outputs without training or supervision.}
    \label{fig:teaser}
\end{center}
}]

\let\thefootnote\relax\footnote{* Equal contribution. † Corresponding authors.

Email: yulong.zhang.longzhao@gmail.com; \quad lgshen@sjtu.edu.cn}

\begin{abstract}
Optical Character Recognition (OCR) is fundamental to Vision-Language Models (VLMs) and high-quality data generation for LLM training. 
Yet, despite progress in average OCR accuracy, state-of-the-art VLMs still struggle with detecting sample-level errors and lack effective unsupervised quality control.
We introduce Consensus Entropy (CE), a training-free, model-agnostic metric that estimates output reliability by measuring inter-model agreement entropy.
 The core insight is that correct predictions converge in output space, while errors diverge. 
 Based on CE, we develop CE-OCR, a lightweight multi-model framework that verifies outputs by ensemble agreement, selects the best outputs, and further improves efficiency through adaptive routing. 
Experiments demonstrate that CE is robust for quality verification, improving F1 scores by 42.1\% over VLM-as-Judge. CE-OCR achieves consistent OCR gains, outperforming self-consistency and single-model baselines at the same cost. Notably, CE requires no training or supervision, enabling plug-and-play integration. Code: \url{https://github.com/Aslan-yulong/consensus-entropy}.
\end{abstract}    
\section{Introduction}

The Optical Character Recognition (OCR) task serves as a critical bridge connecting textual information in the physical world with digital information systems. Therefore, the quality and diversity of OCR-derived data crucially determine the reliability and generalization capability of large language models (LLMs), as it constitutes one of the most important pipelines for acquiring real-world textual data to fuel LLM training processes~\cite{minerU,opendatalab,olmocr}. In recent years, the rapid progress of Vision-Language Models (VLMs)~\cite{liu2023llava,chen2024expanding,gao2024mini,chen2024far,Qwen2.5-VL,Qwen2-VL,glm2024chatglm} has transformed OCR capabilities from task-specific algorithms into a core component of multimodal perceptual understanding. Consequently, OCR accuracy has become a critical metric for assessing the visual-linguistic reasoning capacity of such models. At the same time, the scarcity of high-quality OCR data and the high cost of manual annotation have made both the evaluation of data quality and the assessment of model outputs enduring challenges in OCR research.

Current OCR evaluation methods predominantly rely on standardized benchmark metrics~\cite{OCRBench}, creating a disconnect between reported performance and real-world reliability. Our extensive OCR experiments with state-of-the-art models, including Qwen2.5-VL-72B~\cite{Qwen2.5-VL} and GPT-4o~\cite{openai2024gpt4o}, reveal that even top-performing VLMs exhibit \textbf{frequent semantic inaccuracies and formatting inconsistencies that evade detection by conventional metrics}. Notably, models with higher benchmark scores sometimes underperform in practical applications compared to those with slightly lower benchmark rankings. Current evaluation approaches fall into two categories with distinct limitations: multimodal re-evaluation based on cross-validation is constrained by the evaluation models' own uncertainties and often introduces secondary noise~\cite{zheng2023judging,zhu2023judgelm,kim2023prometheus,gu2024survey,chang2024survey}, while VLM-as-Judge methods for textual quality assessment cannot effectively verify the consistency between visual inputs and textual outputs.
These evaluation challenges hamper trustworthy OCR development for real-world applications. This raises a critical question:
\textit{Since even state-of-the-art VLMs cannot guarantee error-free OCR results, can we leverage models' own capabilities to self-verify and self-improve OCR results without human supervision?}

To answer this question,
we study the behavior of 210 VLMs on OCRBench~\cite{OCRBench} and observe some simple yet powerful patterns: (1) OCR tasks generally admit a unique semantic ground truth, even if formatting may vary; (2) when VLMs predict correctly, their outputs tightly cluster in semantic space; and (3) when they err, the predictions diverge and show high entropy. These insights lead to the introduction of Consensus Entropy (CE), a label-free uncertainty metric that estimates prediction reliability by measuring the entropy of pairwise similarities among model outputs. Low CE indicates high agreement and reliability; high CE signals ambiguity and potential error. CE improves quality verification F1 scores by 15.2, a 42.1\% gain over VLM-as-Judge. Building upon this metric, we develop CE-Ensemble, which uses CE-weighted aggregation to combine multiple model outputs. Finally, we construct the complete CE-OCR framework that incorporates adaptive routing: inputs with low CE are processed by CE-Ensemble, while high-CE cases are routed to stronger VLMs. Experiments on OCRBench, OCRBench-V2~\cite{OCRBench_v2}, and CCOCR~\cite{yang2024ccocr} show that CE-OCR achieves an 8.2\% gain in OCR tasks while only routing 7.3\% samples.

\textbf{Our contributions are three-fold:}
\begin{itemize}
  
    \item We observe that when multiple VLMs process the same image, correct predictions converge while errors diverge—a fundamental behavior not captured by existing OCR methods. 
    This insight motivates \textbf{Consensus Entropy (CE)}, the core contribution of this work. CE is a simple unsupervised metric that identifies trustworthy OCR outputs without labels or retraining, resolving the long-standing challenge of OCR self-verification.
    
    \item We propose a training-free CE-Ensemble and CE-OCR framework with adaptive routing, enabling automatic quality-aware OCR that selectively combines multiple models' strengths or redirects to stronger models for challenging cases. 
    This approach improves OCR accuracy while keeping computational costs nearly flat.
    
    \item We validate CE across diverse OCR tasks and VLMs, demonstrating its effectiveness in output evaluation, data selection, and performance enhancement at low cost.
\end{itemize}
\section{Related Work}

\textbf{Evaluating OCR Capabilities of Multimodal Large Models.} 
OCR has evolved from early commercial tools~\cite{mori1992historical} to modern approaches divided into pipeline systems~\cite{wang2024mineruopensourcesolutionprecise,GROBID,shen2022vila} and end-to-end models~\cite{Nougat,got2}. Recent multimodal large models have demonstrated impressive OCR capabilities, evaluated through both module-level benchmarks~\cite{OCRBench, OCRBench_v2, yang2024ccocr, PubLayNet, fox, Nougat, got2}. However, these benchmarks lack data diversity and comprehensive evaluation. System-level approaches like olmOCR~\cite{olmocr} and MinerU~\cite{he2024opendatalab,wang2024mineruopensourcesolutionprecise} enhance OCR and document understanding, but a key limitation remains: existing frameworks mainly emphasize global accuracy, overlooking the reliability of individual outputs—a critical issue when OCR results feed downstream models~\cite{dai2025careful,dai2025evinote}. Similarly, these models have also been extended to cross-modal generation tasks including text-to-image~\cite{liang2025textcengen}, video-audio~\cite{openmoss_mova_2026}, and text-to-3D~\cite{xiang2025sel3dcraft}, highlighting the potential for broader applications. Our proposed Consensus Entropy metric directly addresses this gap by providing a \textbf{training-free approach to estimate output reliability at the instance level.}

\noindent\textbf{Limitations of LLM-Based Evaluation.} 
OCR evaluation currently relies primarily on the “LLM-as-Judge” paradigm~\cite{zheng2023judging,zhu2023judgelm,kim2023prometheus}, where LLMs serve as automated evaluators. However, comprehensive surveys~\cite{gu2024survey,chang2024survey} identify systematic weaknesses in this approach, including sensitivity to prompt~\cite{liu2024calibrating}, training data contamination~\cite{srivastava2022beyond}, and format biases~\cite{li2023evaluating}. A particularly concerning issue is preference leakage~\cite{li2025preferenceleakagecontaminationproblem}, where correlations between generation and evaluation models inflate evaluation scores. These limitations become more pronounced in adversarial settings~\cite{rando2025adversarialmlproblemsgetting,beyer2025llmsafetyevaluationslackrobustness,dai2026sead}, leading to inconsistent assessments. Traditional metrics like BLEU~\cite{papineni2002bleu} and ROUGE~\cite{lin-2004-rouge} similarly fail to capture semantic accuracy in OCR tasks. Moreover, LLM evaluators have difficulty with multimodal tasks demanding precise layout understanding~\cite{huang2022layoutlmv3}, a capability essential for OCR involving complex layouts and mathematical content. To address this, our CE \textbf{overcomes these limitations by providing model-agnostic quality} assessment without reliance on potentially biased evaluators.

\noindent\textbf{Uncertainty Estimation \& Model Agreement.}
Uncertainty Estimation employs probabilistic methods (modeling parameter uncertainty)~\cite{gal2016dropout,kendall2017uncertainties,liu2025uncertain}, token-level confidence estimation~\cite{DBLP:journals/corr/abs-2307-10236,DBLP:conf/emnlp/ChenDBQWCW23,DBLP:conf/emnlp/ManakulLG23}, and latent space techniques~\cite{wang2024latent,petersen2024uncertaintyquantificationstabledistribution}; cross-modal tasks leverage semantic entropy~\cite{zhang2024vluncertaintydetectinghallucinationlarge} and calibrated confidence~\cite{chen2025enhancing,xiong2024can}, as well as consistency-based verification for structured reasoning~\cite{zhang2025ncvnodewiseconsistencyverification}. Multi-model consensus uses ensemble strategies: selecting outputs~\cite{li2024more,guhasmoothie,si2023getting} or regenerating from selected candidates~\cite{jiang2023llm,tekin2024llm,lv2024urg,chen2025harnessingmultiplelargelanguage}. However, current work focuses on general text generation~\cite{wang2024latent,xiong2024can}, largely overlooking OCR. 
Inspired by prior methods, our CE-OCR framework leverages character-level consensus \textbf{directly from the outputs of any model (open-source or proprietary), without needing access to interior parameters}~\cite{dai2025evinote}.
\section{Method}
\label{sec:method}

Our framework, illustrated in Figure~\ref{fig:teaser}, addresses the fundamental challenge of uncertainty estimation in OCR tasks through a novel multi-model agreement approach. The proposed Consensus Entropy Metric \textbf{transforms quality assessment from a supervised evaluation task into an unsupervised agreement analysis}. 

\begin{figure}[htpt]
  \vspace{-0.8em}
  \centering
  \includegraphics[width=1.0\linewidth]{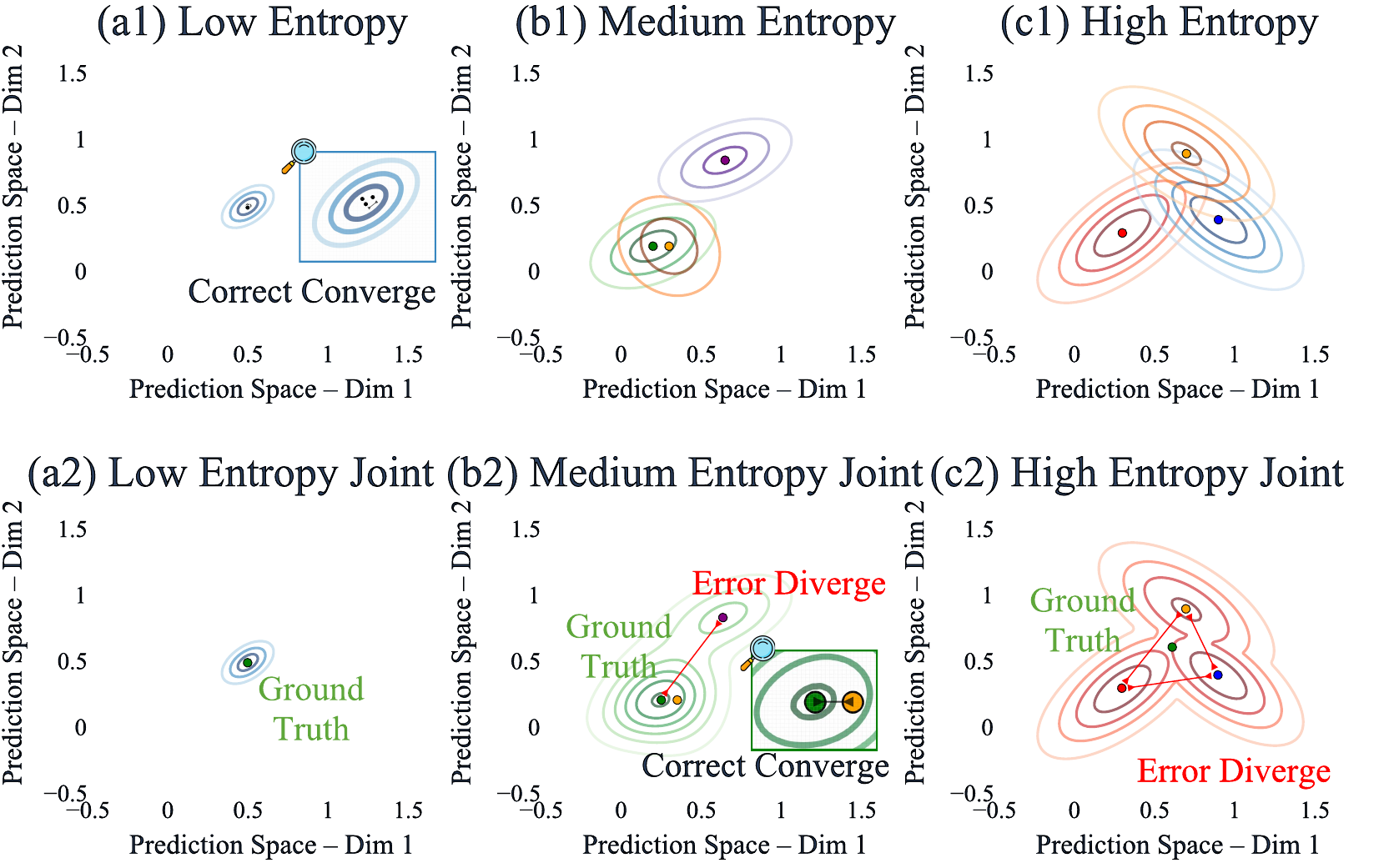}
  \vspace{-0.8em}
  \caption{\textbf{Prediction behaviors across entropy levels.}
  Each plot visualizes VLM predictions in a 2D space. In low-entropy cases (a), predictions tightly cluster around the ground truth (green), while in medium (b) and high-entropy (c) settings, predictions increasingly diverge (Details in Appendix B).}
  \label{fig:entropy}
  \vspace{-1.0em}
\end{figure}

\subsection{Consensus Entropy}
We introduce Consensus Entropy (CE), a novel approach to OCR quality assessment that requires no ground truth supervision. Our method builds on a key insight illustrated in Figure~\ref{fig:entropy}: when multiple independent models process the same image, \textbf{correct outputs naturally converge in a shared representation space while erroneous outputs diverge}. This pattern of convergence and divergence serves as a powerful signal for estimating output quality without requiring labeled data.

Our framework begins by collecting outputs from a set of independent OCR models $\mathcal{M} = \{M_1, M_2, ..., M_n\}$ for an image $I$. To estimate prediction agreement, we measure the pairwise similarity between all model outputs $(O_i, O_j)$. The choice of metric is task-dependent: for tasks requiring character-level precision (e.g., standard OCR, mathematical expressions), we use Edit Distance; for tasks where semantic meaning is key, we compute cosine similarity between output embeddings from a pre-trained text encoder. 

\noindent\textbf{Probability Distribution Derivation.} For character-level tasks, we compute normalized similarity at each position $k$:
\begin{equation}
s_{ij}(k) = 1 - \frac{\text{EditDist}(o_i^k, o_j^k)}{\max(|o_i^k|, |o_j^k|}),
\end{equation}
where $o_i^k$ and $o_j^k$ are subsequences up to position $k$. The probability distribution is then:
$p_{ij}(k) = {s_{ij}(k)}/{\sum_{j'} s_{ij'}(k)}$.
For example, given three VLM outputs for an invoice number---``Invoice'', ``1nvoice'', ``Invoice''---the normalized similarities to the first output are $(1.0, 0.86, 1.0)$, yielding $p = (0.35, 0.30, 0.35)$ and entropy $H = 1.09$ (near-uniform), whereas two identical outputs would yield $H = 0$ (perfect agreement).

We then calculate the pairwise entropy $\mathcal{E}$ between all model output pairs:
\begin{equation}
\mathcal{E}_{ij} = -\sum_{k} p_{ij}(k) \log p_{ij}(k),
\end{equation}
where $p_{ij}(k)$ is the probability distribution derived from the similarity between the $i$-th and $j$-th model outputs at position $k$. Unlike using a single scalar value like cosine similarity for the entire output, our entropy-based approach captures the distributional uncertainty between outputs.

For each model output, we then compute its average entropy distance to all other outputs:
$\overline{\mathcal{E}}_{i} = \frac{1}{n-1} \sum_{j \neq i} \mathcal{E}_{ij}$.
This average entropy distance $\overline{\mathcal{E}}_{i}$ estimates how consistent each model's output is with the collective predictions—lower values indicate outputs that align well with the consensus.

To derive the final Consensus Entropy score $\delta$, we model the overall outputs distribution. For semantic tasks, we use \textbf{kernel density estimation (KDE)}~\cite{silverman2018density} on the output embeddings in output space $\Omega$, where each output's contribution is weighted inversely by its average entropy distance $\overline{\mathcal{E}}_{i}$. For tasks using edit distance, where outputs do not have an explicit vector representation, we compute the final entropy from the distribution of pairwise distances. Both approaches yield a final Consensus Entropy value $\delta$.

\begin{equation}
p(\mathbf{v}) = \sum_{i=1}^n w_i \cdot \mathcal{N}(\mathbf{v} | \mathbf{v}_i, \Sigma_i),
\end{equation}

where $w_i$ is inversely proportional to $\overline{\mathcal{E}}_{i}$, giving greater weight to outputs with lower entropy distance (higher consensus), and $\mathcal{N}(\mathbf{v} | \mathbf{v}_i, \Sigma_i)$ is a multivariate Gaussian kernel centered at $\mathbf{v}_i$ with covariance matrix $\Sigma_i$ that adapts based on the local density of predictions.

In practice, for the semantic case, we can discretize the semantic space using an $N \times N$ grid. The probability at each cell is computed based on the weighted KDE, and the final Consensus Entropy $
\delta = -\sum_{i=1}^{N^2} p_i \log p_i$, 
where $p_i$ is the probability at cell $i$ in the grid. This discretization enables efficient computation while maintaining accuracy.

\begin{figure}
  \centering
  \includegraphics[width=0.9\linewidth]{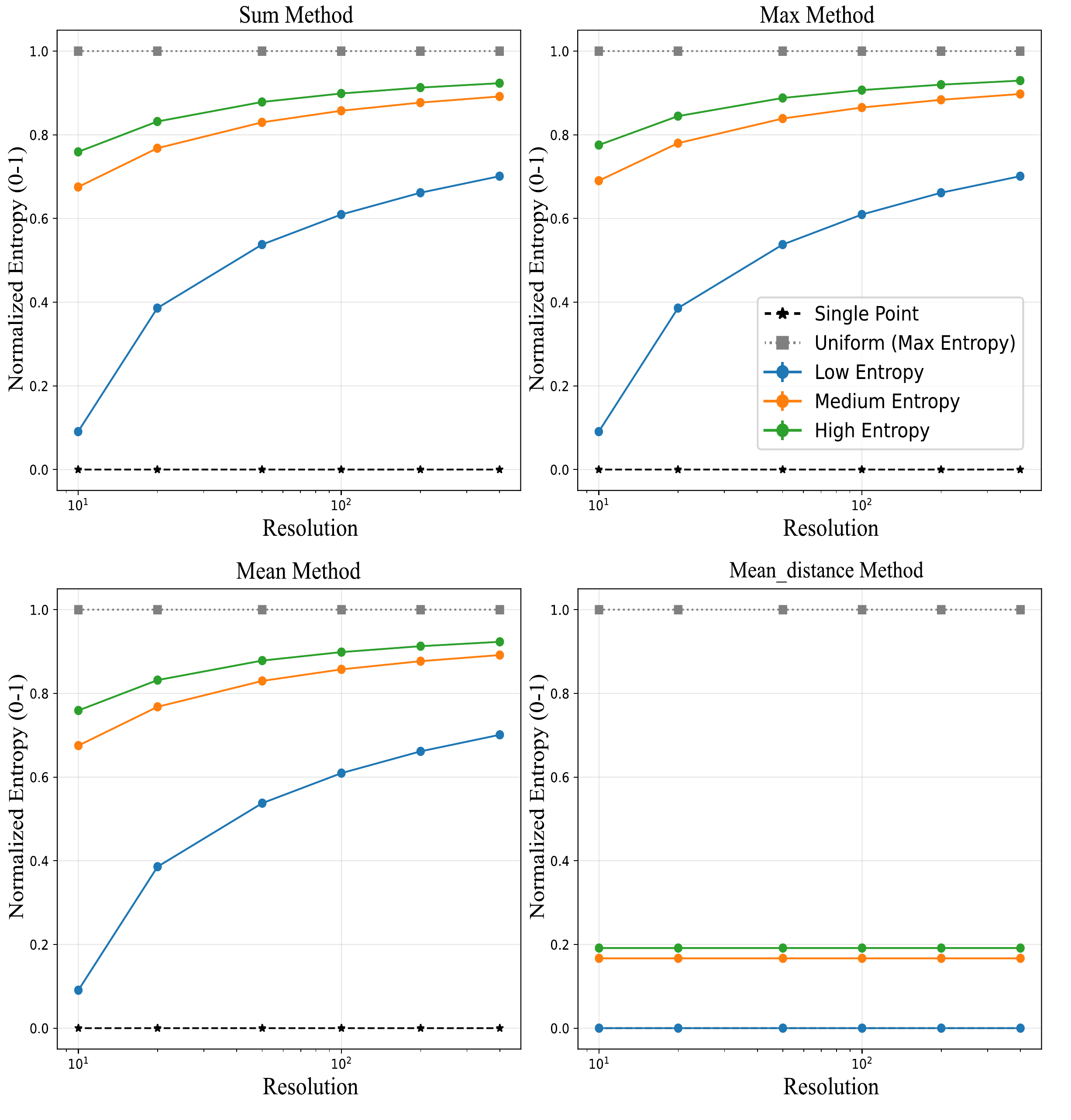}
  \vspace{-1.0em}
  \caption{Normalized entropy [0,1] analysis of four combination methods. Single point (dashed) and uniform (dotted) distributions as bounds.}
  \label{fig:norm_entropy_method}
  \vspace{-1.3em}
\end{figure}

To identify the optimal strategy for combining probability distributions in our CE framework, we evaluated four aggregation methods: Sum, Max, Mean, and Mean Distance. As shown in Figure~\ref{fig:norm_entropy_method}, all methods maintain consistent preference ordering, indicating robustness of the entropy-based approach. The Mean Distance method directly averages the $\overline{\mathcal{E}}$ values and is \textbf{grid-invariant} (CE values remain consistent across varying resolutions $N$) and \textbf{order-preserving} (higher agreement always yields lower CE). These properties, validated by simulations (Appendix B), make Mean Distance the most robust choice for practical deployment. Additional analyses are provided in Appendix B.

As shown in Figure~\ref{fig:entropy}, this entropy value effectively captures \textbf{the degree of agreement among model predictions}. In low-entropy scenarios (Figure~\ref{fig:entropy}: a2), predictions form tight clusters around the ground truth. In contrast, medium-entropy (Figure~\ref{fig:entropy}: b2) and high-entropy (Figure~\ref{fig:entropy}: c2) settings show progressively greater divergence.

The Consensus Entropy $\delta$ serves as our core uncertainty metric—lower values indicate higher agreement and likely correctness, while higher values signal potential errors or ambiguities. This metric provides a direct measure of output reliability \textbf{without requiring ground truth labels, enabling fully unsupervised quality assessment at scale}. As referenced in Figure~\ref{fig:teaser}, $\delta$ serves as the key decision signal in our routing framework, determining whether to accept ensemble predictions or seek expert refinement.

\subsection{Ensemble and Router}

The CE framework is designed to be flexible and adaptable, allowing for various ensemble strategies and routing mechanisms based on the computed Consensus Entropy. The core idea is to \textbf{leverage the entropy value to guide the decision-making process} for handling OCR outputs, balancing between efficiency and quality.

\noindent\textbf{Threshold Gate and Routing Mechanism.} After computing Consensus Entropy $\delta$, we introduce a \textbf{threshold-based routing mechanism to balance efficiency and quality}. The key component is a threshold gate parameter $\theta$ that determines how outputs are processed:

\begin{equation}
  R(\delta, \theta) =
  \begin{cases}
  0, & \text{if } \delta \leq \theta \\
  1, & \text{if } \delta > \theta
  \end{cases}
\end{equation}
  
  where $R(\delta, \theta) = 0$ indicates using the ensemble output, and $R(\delta, \theta) = 1$ triggers routing to a stronger VLM. $\theta$ serves as a critical decision boundary that can be adjusted based on task-specific requirements, as illustrated in Figure~\ref{fig:teaser}. Lower thresholds prioritize quality by routing more samples to expert models, while higher thresholds favor computational efficiency by accepting more ensemble predictions.
We calibrate $\theta$ through empirical analysis on development data, identifying the entropy value that optimizes the quality-efficiency trade-off. Our experiments \textbf{indicate that $\theta$ values around 0.5} provide an optimal balance for most OCR tasks, effectively identifying high-quality predictions while routing only the most problematic samples for expert processing.

\noindent\textbf{Ensemble Strategy.} When Consensus Entropy falls below the threshold ($\delta \leq \theta$), we apply a weighted ensemble strategy to combine predictions from multiple models. Unlike simple averaging approaches, our weights directly leverage the entropy framework:
$O_{\text{ens}} = \sum_{i=1}^{n} w_i \cdot O_i$,
where the weighting coefficients $w_i$ are derived from the average entropy distances $\overline{\mathcal{E}}_i$:
\begin{equation}
w_i = \frac{1/\overline{\mathcal{E}}_{i}}{\sum_{j=1}^{n} 1/\overline{\mathcal{E}}_{j}}.
\end{equation}

This approach assigns \textbf{higher weights to outputs with lower entropy distances} (higher consensus), while \textbf{downweighting outputs with higher entropy distances} (potential outliers).
In practice, implementing this weighted combination for text requires a token-level aggregation approach. We first align all outputs using \textbf{dynamic programming to identify corresponding tokens across predictions}. Then, for each position, we select the token with the highest weighted consensus
$t_k^* = \underset{t \in T_k}{\text{argmax}} \sum_{i:t \in O_i} w_i$,
where $T_k$ is the set of all candidate tokens at position $k$ across model outputs, and the sum aggregates weights from all models whose outputs contain token $t$ at that position.

\noindent\textbf{Expert Routing.} When CE exceeds the threshold ($\delta > \theta$), indicating low agreement among models, we route the sample to a more powerful VLM (referred to as $M_{\text{exp}}$) for rephrasing. This stronger model processes the input image along with the ensemble prediction and original model outputs as context:

\begin{equation}
O_{\text{final}} = M_{\text{exp}}(I, \{O_1, O_2, \ldots, O_n\}, O_{\text{ens}}).
\end{equation}

The framework enables a dynamic, quality-aware OCR pipeline that automatically adapts to input complexity. By leveraging CE to guide both ensemble aggregation and expert routing decisions, \textbf{we achieve superior OCR performance without requiring additional training or supervision.} 
Complete algorithms are presented in Appendix E. 

\section{Experiments}

\begin{table*}[t]
  \centering
  \caption{\textbf{OCR-Verifying: F1 comparison of VLM-as-Judge (VLM-J) vs.\ CE on Human-Annotated Data.} For fair comparison, both methods use a \textbf{single} reference VLM (no ensemble): VLM-J inputs image + text to a reference model for scoring; CE generates OCR with the same reference model and computes consensus entropy between two sequences.}
  \vspace{-0.6em}
  \begin{tabular}{c|cc|cc|cc}
  \hline
  \textbf{Human Score}
        & \multicolumn{2}{c|}{\textbf{GPT4o}} & \multicolumn{2}{c|}{\textbf{Qwen2-VL-7B}} & \multicolumn{2}{c}{\textbf{Qwen2-VL-72B}} \\
  \textbf{Band} & VLM-J & CE (Ours) & VLM-J & CE (Ours) & VLM-J & CE (Ours) \\
  \hline
  0.9-1.0 & \textbf{72.12} & 57.29 & 70.05 & \textbf{70.88} & \textbf{72.27} & 66.13 \\
  0.7-0.8 & 18.42 & \textbf{40.19} & 19.82 & \textbf{39.56} & 23.01 & \textbf{40.14} \\
  0.4-0.6 & 21.47 & \textbf{38.67} & 0.00 & \textbf{27.34} & 11.43 & \textbf{28.47} \\
  0.0-0.3 & 48.21 & \textbf{65.44} & 54.74 & \textbf{67.42} & 52.83 & \textbf{69.57}   \\
  \hline
  \textbf{Overall} & 40.0 & \textbf{48.0 (+20.0\%)} & 36.1 & \textbf{51.3 (+42.1\%)} & 39.8 & \textbf{51.0 (+28.1\%)} \\
  \hline
  \end{tabular}
  \label{tab:filtered_eval_table}
  \vspace{-1.0em}
  \end{table*}

We evaluate CE through: (1) unsupervised quality verification, using CE as a standalone filter to identify correct vs.\ incorrect outputs without any ensemble or routing; (2) CE-Ensemble, which incorporates CE-weighted multi-model aggregation for performance gains; and (3) full CE-OCR, which combines CE-Ensemble with adaptive routing and stronger-model rephrasing, followed by ablation studies to isolate each component's contribution and beyond-OCR exploration to demonstrate broader applicability.
We structure our evaluation progressively: this section (\S4.2) validates CE alone as a verification metric; \S4.3 introduces CE-Ensemble by adding multi-model aggregation; \S4.4 extends to the full CE-OCR pipeline with routing and rephrasing.

\subsection{Experimental Setup}

\textbf{Implementation.} We use inference-only evaluation with diverse VLMs following \textit{vlmevalkit} \cite{vlmevalkit} defaults. Most models run on single NVIDIA H800 (80GB) GPU; largest models (e.g., Qwen2.5-VL-72B) use 2-4 GPUs. Our work involves \textbf{only inference, eliminating training and minimizing computational cost.}
Edit distance uses Levenshtein; cosine uses bge-m3~\cite{bge-m3}, applied only to OCRBench-V2.
All experiments apply temperature 0.0 (single-run) except Self-Consistency (T=0.7, 3 runs).

\noindent\textbf{Datasets.} Primary evaluation on OCRBench~\cite{OCRBench}, OCRBench-V2~\cite{OCRBench_v2} and CCOCR~\cite{yang2024ccocr}. For human alignment assessment, we curate 1K PDF pages from real-world scenarios, manually scored (1.0: Perfect to 0.0: Mostly Incorrect) based on text matching accuracy. We report all scores on a scale of 0.0-100.0, except OCRBench (0-1000).

\noindent\textbf{Baselines.}
We use \textbf{VLM-as-Judge}, employing GPT-4o~\cite{openai2024gpt4o}, Qwen2-VL-72B, and Qwen2-VL-7B~\cite{Qwen2-VL} with OCR-quality prompts; \textbf{VL-Uncertainty}~\cite{zhang2024vluncertaintydetectinghallucinationlarge}, which measures hallucination via semantic clustering on LLM embeddings; \textbf{Single Model}, directly outputting VLM predictions; \textbf{Self-Consistency}~\cite{wang2022self}, selecting by voting among multiple samples from the same VLM; and \textbf{ROVER}~\cite{fiscus1997post}, a classical ensemble method using discrete voting on text outputs.

\subsection{Unsupervised OCR Self-Verification with CE}

\textbf{Setup.} Comprehensive experiments are conducted on OCRBench across varying CE thresholds.
We also evaluate the unsupervised identification capability of CE for accurate OCR predictions against VLM-as-Judge baselines.

\begin{figure}
  \centering
  \includegraphics[width=1.0\columnwidth]{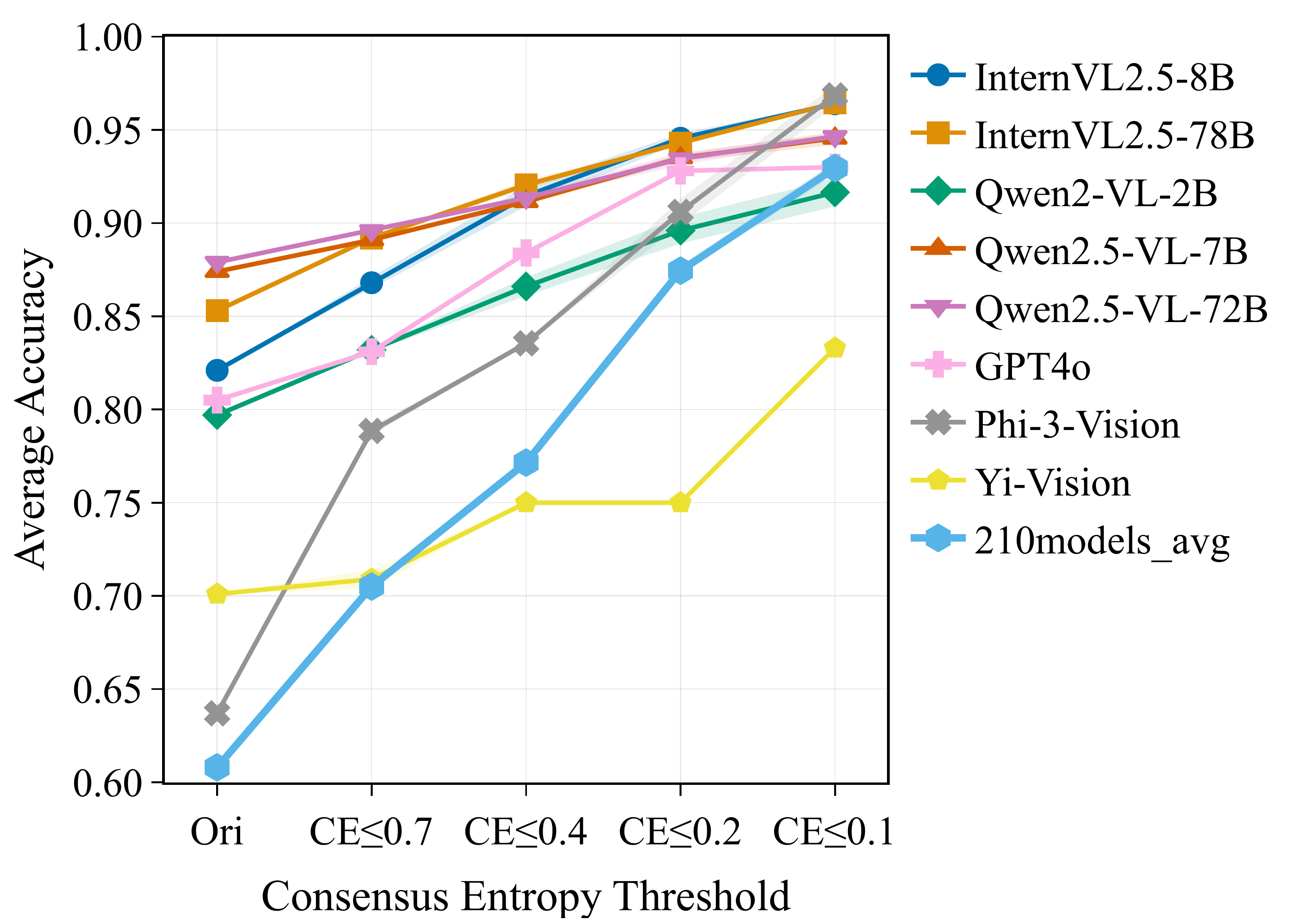}
  \vspace{-1.8em}
  \caption{\textbf{Model Performance on OCRBench under Different CE Thresholds.} A pure filtering experiment: each target model is paired with one reference model (ref1: Qwen2VL-7B; ref2: Qwen2VL-72B) to compute CE---no ensemble or rephrase is applied. 210models\_avg: average across 210 models. Shaded area covers ref1 and ref2 accuracy; solid line is their mean.}
  \label{fig:ocr_th}
  \vspace{-1.0em}
\end{figure}

\noindent\textbf{CE Threshold Analysis.} Figure~\ref{fig:ocr_th} illustrates the relationship between CE thresholds and model accuracy across different VLMs. The results show a clear trend: as the CE threshold decreases, the average accuracy of model responses increases. \textbf{This pattern holds true for the majority of the 210 VLMs tested, including both open-source and proprietary models.}

\textbf{Human Evaluation Comparison.} Utilizing our curated dataset of 1,000 PDF pages from real-world scenarios, which includes documents of various languages, types, and complexities, we compared the CE method with the VLM-as-Judge approach. Table~\ref{tab:filtered_eval_table} shows that CE method consistently achieves higher F1 scores, ranging from 48.0 to 51.3, compared to 36.1 to 40.0 for the VLM-as-Judge. This marks an average improvement of 15.2 in F1 scores for OCR quality verification, a 42.1\% gain over baselines. The robustness of CE across various document complexities highlights its effectiveness for practical OCR applications.

\subsection{CE-Ensemble Evaluation}

\textbf{Setup.} We evaluate CE-Ensemble on OCRBench using 24 models, testing all 3-5 model combinations ($\mathrm{C}_{24}^{3}
=2024$, $\mathrm{C}_{24}^{4}=10626$, $\mathrm{C}_{24}^{5}=42504$, 
55154 combinations in total, full results in Appendix I ).

\begin{table}[htbp]
\vspace{-0.5em}
  \centering
  \caption{\textbf{Statistical analysis of performance gains from CE-Ensemble outputs.} The table summarizes average improvements ($\Delta_{\max}^{+}$: average gain of ensemble over the lowest single-model score, $\Delta_{\min}^{+}$: over the best single-model score, $\Delta_{\text{avg}}^{+}$: over the average score of all participating models) and the proportion of positive cases across 3-5 model ensembles using CE-based output selection on OCRBench. CE yields positive gains in most cases.}
  \vspace{-0.5em}
  \label{tab:aggregation_delta_stats}
  \setlength{\tabcolsep}{3pt}
  \begin{tabular}{c|ccc|ccc}
  \hline
  Size & $\Delta_{\max}^{+}$ & $\Delta_{\min}^{+}$ & $\Delta_{\text{avg}}^{+}$ & $\Delta_{\max}^{+\%}$ & $\Delta_{\min}^{+\%}$ & $\Delta_{\text{avg}}^{+\%}$ \\
  \hline
  3 & 50.2 & 4.2 & 28.3 & 100\% & 66.2\% & 94.7\% \\
  4 & 67.5 & 12.7 & 41.7 & 100\% & 82.2\% & 100.0\% \\
  5 & 78.3 & 17.8 & 50.3 & 100\% & 91.11\% & 100.0\% \\
  \hline
  \end{tabular}
  \vspace{-0.5em}
\end{table}

\noindent\textbf{Statistical analysis.}
Table~\ref{tab:aggregation_delta_stats} shows that \textbf{as ensemble size increases, both the average and minimum gains improve significantly}. Notably, $\Delta_{\max}^{+\%}$ remains at 100\% for all sizes, indicating CE always avoids the worst predictions. Meanwhile, the steady rise of $\Delta_{\min}^{+\%}$ and the near-100\% $\Delta_{\text{avg}}^{+\%}$ demonstrate that CE-Ensembles increasingly select predictions close to the best in each group. This confirms that CE is robust and effective in leveraging multiple models to approach or surpass the best single-model performance.

\noindent\textbf{Representative Ensembles Analysis.}  
We select top VLMs and well-known open-source VLMs as Examples in Table~\ref{tab:model_combinations}. Table~\ref{tab:model_combinations} shows that CE ensembles not only \textbf{surpass the current state-of-the-art (SOTA) score of 926}, but also \textbf{leverage weaker or smaller models to further boost the performance ceiling of SOTA models.} Notably, ensembles \textbf{enable low-cost open-source models (with fewer than 10B parameters) to outperform much larger high-cost closed-source SOTA models}, highlighting the effectiveness of CE in enabling resource-efficient models to exceed the performance of high-cost, large-scale alternatives.

\begin{table}[htbp]
\vspace{-0.5em}
  \centering
  \caption{\textbf{Representative CE Ensemble Examples On OCRBench.}
  $\dagger$: cheap open-source models ($<10$B) outperform larger closed-source SOTA models. $*$: weaker models further boost the upper limit of SOTA models.}
  \vspace{-0.8em}
  \label{tab:model_combinations}
  \small
  \begin{tabular}{l}
  \toprule
  \textbf{Model Combinations of CE-Ensemble (Individual Scores)} \\
  \midrule
  $^*$Ovis2-1B, Qwen2.5VL-7B, Step1V, \textbf{Step1o} \\
  (890, 874, 886, \textbf{926}) \hfill \textbf{CE-Ensemble: 955 \qquad Gain: +29} \\
  \midrule
  $^\dagger$Ovis2-1B, \textbf{Ovis2-4B}, Qwen2VL-7B, Qwen2.5VL-7B \\
  (890, \textbf{909}, 843, 874) \hfill \textbf{CE-Ensemble: 933 \qquad Gain: +24} \\
  \midrule
  $^*$Ovis2-4B, Qwen2.5VL-7B, \textbf{Step1o} \\
  (909, 874, \textbf{926}) \hfill \textbf{CE-Ensemble: 938 \qquad Gain: +12} \\
  \midrule
  InternVL2.5-78B, Qwen2.5VL-72B, \textbf{Qwen2VL-72B} \\
  (853, 879, \textbf{888}) \hfill \textbf{CE-Ensemble: 920 \qquad Gain: +32} \\
  \midrule
  $^\dagger$InternVL2.5-8B, Qwen2VL-7B, \textbf{Qwen2.5VL-7B} \\
  (821, 843, \textbf{874}) \hfill  \textbf{CE-Ensemble: 897 \qquad Gain: +23} \\
  \bottomrule
  \end{tabular}
  \vspace{-1.5em}
  \end{table}

\subsection{Evaluation on CE-OCR for Self-Improving}

\textbf{Setup.} Experiments are conducted on multiple benchmarks (threshold=0.5). Below-threshold samples use weighted ensemble, above-threshold samples route to stronger VLM.

\begin{table}[htbp]
\vspace{-1em}
\centering
\caption{\textbf{Performance comparison of CE-OCR and CE-Ensemble (Cosine Distance) against individual models and basic ensemble on OCRBench-V2 task categories. }
En: English OCR, Elem: Element Parsing, Cn All: Chinese Overall.
}
\vspace{-0.8em}
\label{tab:ce_framework_tasks}
\small
\setlength{\tabcolsep}{3pt}
\begin{tabular}{lcccc}
\toprule
\textbf{Method} & \textbf{En} & \textbf{Math} & \textbf{Elem} & \textbf{Cn All} \\
\midrule
GPT4o & 61.2 & 43.4 & 29.8 & 32.2 \\
InternVL2.5-26B & 65.6 & 37.4 & 32.6 & 44.2 \\
Gemini Pro & 61.2 & 47.7 & 30.9 & 43.1 \\
\midrule
CE-Ensemble & 67.2 & 50.1 & \textbf{34.0} & 45.7 \\
CE-OCR (GPT4o Rephrase) & \textbf{71.6} & \textbf{53.1} & 33.8 & \textbf{48.0} \\
\midrule
Relative $\Delta$ vs Ensemble (\%) & +6.5\% & +6.0\% & -0.5\% & +5.0\% \\
Relative $\Delta$ vs Best Single (\%) & +9.1\% & +11.3\% & +3.7\% & +8.6\% \\
\bottomrule
\end{tabular}
\end{table}

\begin{figure}
  \centering
  \includegraphics[width=1.0\columnwidth]{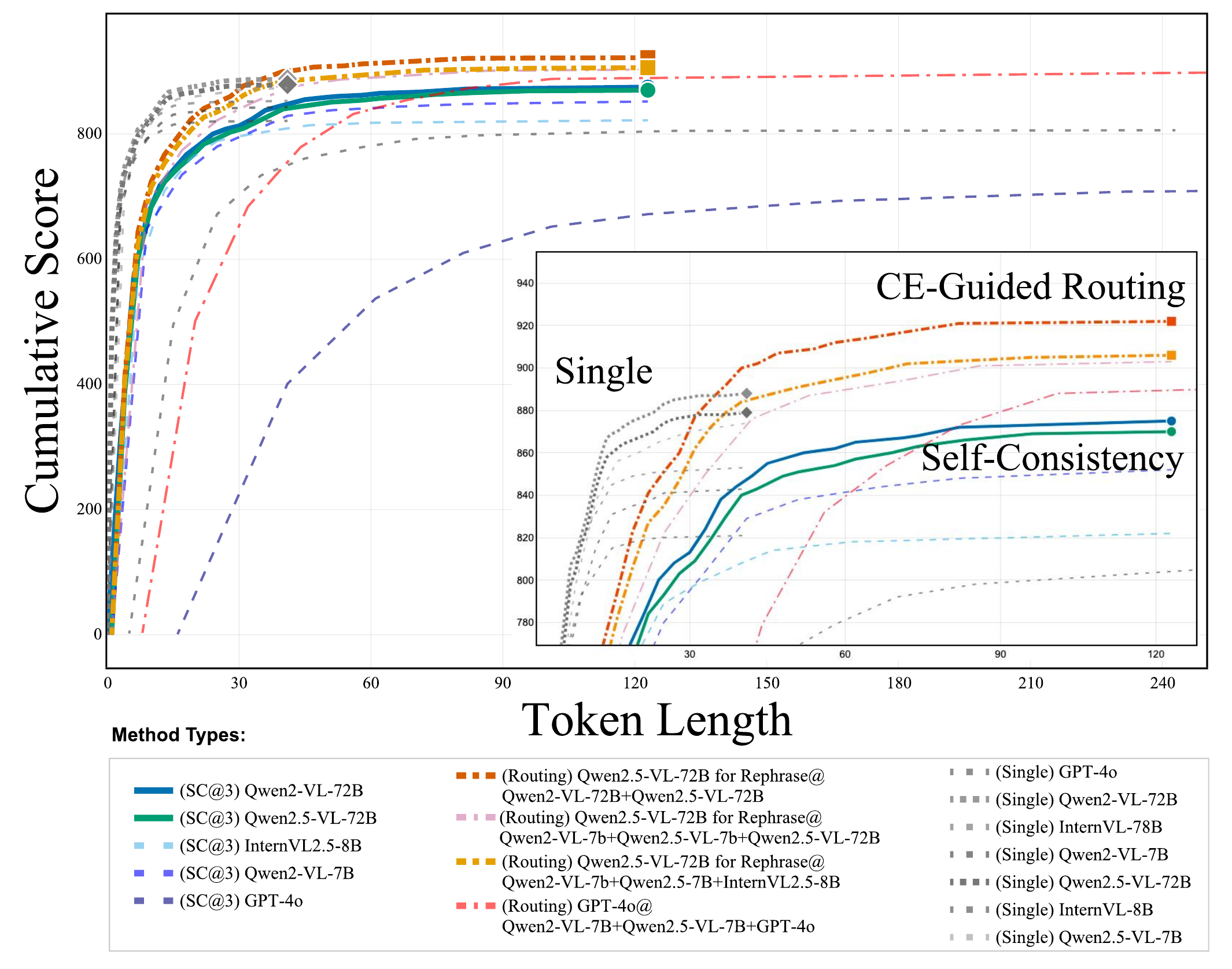}
  \vspace{-1.0em}
  \caption{\textbf{Performance comparison across token lengths on OCRBench.} SC@3: Self-Consistency with 3 samples; Routing: ensemble with rephrasing; Single: Single models.}
  \label{fig:ocr_performance_comparison}
  \vspace{-0.5em}
\end{figure}

\begin{table}[htbp]
    \centering
    \caption{\textbf{Performance comparison of different OCR methods based on cumulative scores.}}
    \vspace{-1.0em}
    \label{tab:ocr_method_comparison}
    \small
    \setlength{\tabcolsep}{2.5pt}
    \begin{tabular}{lcc}
    \toprule
    \textbf{Method} & \textbf{Score} & \textbf{Improvement (\%)} \\
    \midrule
    SC@3 (Average) & 828 & -2.8\%\\
    SC@3 (Best: Qwen2VL-72B) & 875 & +2.7\%\\
    Routing (Average) & \textbf{907} & +6.5\%\\
    Routing (Best: Qwen2.5VL-72B) & \textbf{922} & \textbf{+8.2\%}\\
    Single (Average) & 852 & -\\
    Single (Best: Qwen2VL-72B) & 888 & -\\
    \bottomrule
    \end{tabular}
    \vspace{-1.0em}
\end{table}

\noindent\textbf{Results Analysis.} Table~\ref{tab:ce_framework_tasks} shows that our CE-OCR framework demonstrates superior performance over both individual models and basic ensemble approaches across a broad spectrum of OCR tasks, particularly excelling in complex scenarios such as mathematical calculations, where it achieves a 6.0\% accuracy improvement over conventional ensembles. The framework's efficacy is derived from its dynamic selection mechanism based on CE values. Inputs with high consensus among models leverage weighted ensemble outputs, while those with low consensus are directed to a stronger VLM for improved prediction. This selective routing strategy optimizes resource utilization, with \textbf{only 7.3\% of inputs requiring stronger model rephrasing} (Table~\ref{tab:ablation_components}). 
According to Table~\ref{tab:ocr_method_comparison}, our method substantially outperforms Self-Consistency. While SC averages outputs from identical prompts, our approach dynamically selects processing strategies based on model consensus. Figure~\ref{fig:ocr_performance_comparison} shows our models maintain superior performance across all token lengths, demonstrating \textbf{CE-OCR maintains higher cumulative scores as token length increases.}
 By estimating uncertainty and effectively integrating the strengths of multiple models, the framework offers a practical, training-free solution for diverse real-world OCR challenges.

\subsection{Ablation Studies}

We conduct ablation studies to evaluate each component's contribution. Table~\ref{tab:ablation_components} shows the results when removing individual components from our full framework.

\begin{table}[htbp]
  \centering
  \caption{\textbf{Ablation study results on OCRBench.} We examine the CE Framework with different components removed.}
  \vspace{-0.8em}
  \label{tab:ablation_components}
  \small
  \setlength{\tabcolsep}{2.5pt}
  \begin{tabular}{lccc}
  \toprule
  \textbf{Configuration}  & \textbf{Score↑} & \textbf{\% Routed} & \textbf{Rel. Perf.} \\
  \midrule
  w/o CE (Single Max) & 888 & 0\% & 97.9\% \\
  w/o Ensemble (Single Avg) & 852 & 0\% & 93.9\% \\
  w/o Routing (All Ensemble) & 902 & 0\% & 99.4\% \\
  Full CE-OCR Framework& \textbf{907} & 7.3\% & \textbf{100\%} \\
  \bottomrule
  \end{tabular}
  \vspace{-0.5em}
\end{table}

\noindent\textbf{Consensus Entropy (CE).} Removing CE leads to a performance drop and eliminates the ability to verify output quality, demonstrating its critical role in uncertainty quantification. CE significantly outperforms traditional VLM-as-Judge methods, achieving an F1 score improvement of 15.2 (Table~\ref{tab:filtered_eval_table}). 
This validates CE's capability to assess OCR quality without explicit supervision. 
CE works because higher entropy indicates model disagreement, which correlates with output uncertainty and enables unsupervised quality assessment.

\noindent\textbf{Multi-Model Ensemble.} Using single models instead of the ensemble results in a 7.2\% performance degradation. The ensemble's diversity is essential for optimal CE calculation and robust performance across different input types. CE improves ensemble performance by providing quantifiable comparison values that enable unsupervised selection of the best model outputs.

\noindent\textbf{Adaptive Routing \& Rephrasing.} The framework without routing achieves 99.4\% of full performance but processes all inputs with the ensemble. Our adaptive routing reduces computational cost by routing only 7.3\% of inputs to stronger models while maintaining near-identical accuracy. CE enables unsupervised identification of low-quality outputs, which are then improved through stronger models or better prompt rephrasing.


\begin{table}[htbp]
  \centering
  \caption{\textbf{Threshold sensitivity analysis on OCRBench-V2.} Each cell shows accuracy improvement ($\Delta \times 10^2$) / rephrase percentage (\%). Lower $\theta$ means more samples rephrased.}
  \label{tab:threshold_sensitivity}
  \small
  \setlength{\tabcolsep}{2.0pt}
  \begin{tabular}{l|c|c|c|c|c}
  \toprule
  \textbf{Model} & $\theta=0.95$ & $\theta=0.9$ & $\theta=0.8$ & $\theta=0.6$ & $\theta=0.2$ \\
  \midrule
   Gemini Pro& +1.6/11.5 & +3.1/22.3 & +4.3/43.5 & +5.4/66.3 & +6.1/91.5 \\
   GPT-4o & +3.8/15.0 & +5.1/24.9 & +6.8/48.7 & +7.7/71.3 & +8.8/91.2 \\
  \bottomrule
  \end{tabular}
  \vspace{-0.5em}
\end{table}

\noindent\textbf{Threshold Sensitivity.} We analyze the impact of routing threshold $\theta$ on the accuracy-computation trade-off (Table~\ref{tab:threshold_sensitivity}). Lower $\theta$ values route more samples for rephrasing, improving accuracy but increasing computational cost. The optimal $\theta$ balances performance gains with efficiency.

\noindent\textbf{Model Diversity \& Other Experiments.} We evaluate CE-Ensemble using both identical model instances and models from the same architectural series. Table~\ref{tab:same_family_gains} demonstrates consistent gains across same-family ensembles and identical-model sampling. While diversity enhances CE, it remains effective with architecturally similar models, capturing task uncertainty through sampling variance. 

\begin{table}[htbp]
 \vspace{-0.5em}
  \centering
  \caption{\textbf{Same-family model gains.} 
  Identical row: same model with stochastic sampling (T=0.7); Families: greedy (T=0).}
  \label{tab:same_family_gains}
  \small
  \setlength{\tabcolsep}{3pt}
  \begin{tabular}{l|ccc}
  \toprule
  \textbf{Model Family} & \textbf{Avg Gain} & \textbf{Max Gain} & \textbf{$>$1\% Cases} \\
  \midrule
  InternVL & +1.24\% & +1.7\% & 75\% \\
  Ovis2 & +1.06\% & +1.2\% & 67\% \\
  QwenVL & +1.83\% & +2.9\% & 79\% \\
  Identical (T=0.7, 3 runs) & +2.1\% & +3.2\% & 100\% \\
  \bottomrule
  \end{tabular}
  \vspace{-0.8em}
\end{table} 

\subsection{Comparison with Classical Baselines}

Table~\ref{tab:rover_comparison} compares CE-Ensemble with representative ensemble methods under the same 3-VLM setup. ROVER's discrete word-level voting works for closed-vocabulary ASR but fails catastrophically on open-ended VLM tasks (Doc-VQA $-42.1\%$, Math-VQA $-92.0\%$), because VLM outputs are diverse and non-aligned in phrasing. VL-Uncertainty \cite{zhang2024vluncertaintydetectinghallucinationlarge} uses semantic clustering on LLM embeddings; it achieves modest gains on structured OCR ($+3.3\%$) but degrades on semantic VQA tasks (Doc-VQA $-10.1\%$, KR $-3.5\%$), as character-level differences critical for OCR are invisible to semantic embeddings. In contrast, CE-Ensemble improves across all four categories ($+8.2\%$ average), confirming that output-space agreement analysis is more robust for generation tasks.

\begin{table}[htbp]
\centering
\caption{\textbf{Comparison with other ensemble methods.} Best Single: highest score among participating models. All methods use the same 3-VLM setup for fair comparison. CE consistently outperforms all baselines across tasks.}
\label{tab:rover_comparison}
\setlength{\tabcolsep}{3pt}
\begin{tabular}{l|cccc|c}
\toprule
\textbf{Method} & \textbf{OCR} & \textbf{Doc-VQA} & \textbf{Math} & \textbf{KR} & \textbf{Avg $\Delta$} \\
\midrule
Best Single & 61.2 & 87.5 & 40.0 & 60.3 & 0\% \\
VL-Uncertainty & 64.5 & 78.7 & 45.1 & 58.2 & +0.2\% \\
ROVER & 57.5 & 50.7 & 3.2 & 50.4 & -33.8\% \\
CE-Ensemble & \textbf{67.2} & \textbf{90.5} & \textbf{45.6} & \textbf{66.3} & \textbf{+8.2\%} \\
\bottomrule
\end{tabular}
\vspace{-0.8em}
\end{table}

\subsection{Layout-Sensitive \& VQA Tasks Beyond OCR}

\noindent\textbf{Setup.} We evaluate CE's effectiveness on diverse VQA tasks to verify its sensitivity to layout structure and semantic understanding. We employ Edit Distance (character-level) for layout-sensitive tasks and Cosine Distance (bge-m3~\cite{bge-m3}) for semantic VQA.

\noindent\textbf{Layout-Sensitive Tasks.} Table~\ref{tab:vqa_beyond_ocr} shows Edit Distance-based CE achieves strong gains on layout-heavy tasks: Doc-VQA (+3.4\%), Formula (+7.3\%), Math-VQA (+14.0\%), confirming CE effectively captures spatial structure through character-level divergence.

\noindent\textbf{Semantic VQA.} Cosine Distance-based CE similarly excels on semantic tasks: Scene-VQA (+6.0\%), Science-VQA (+3.9\%), Knowledge-Reasoning (+10.0\%), demonstrating CE's broad applicability beyond OCR.

\begin{table}[htbp]
\vspace{-0.3em}
\centering
\caption{\textbf{CE performance on non-OCR VQA tasks.}  Baseline: highest accuracy scores among participating models;
Better metric: distance metric with greater improvement.}
\vspace{-0.8em}
\label{tab:vqa_beyond_ocr}
\small
\setlength{\tabcolsep}{1.5pt}
\begin{tabular}{lccc}
\toprule
\textbf{Category} & \textbf{Baseline} & \textbf{CE-Ensemble (\(\Delta\))} & \textbf{Better Metric} \\
\midrule
Scene-VQA (easy) & 92.5 & \textbf{98.0 (+6.0\%)} & Cosine Distance \\
Doc-VQA (easy) & 87.5 & \textbf{90.5 (+3.4\%)} & Edit Distance \\
Formula (easy) & 82.0 & \textbf{88.0 (+7.3\%)} & Edit Distance \\
Science-VQA & 61.3 & \textbf{63.7 (+3.9\%)} & Cosine Distance \\
Math-VQA & 40.0 & \textbf{45.6 (+14.0\%)} & Cosine Distance \\
Knowl.-Reason. & 60.3 & \textbf{66.3 (+10.0\%)} & Cosine Distance \\
Visual-Und. & 75.7 & \textbf{82.4 (+8.9\%)} & Cosine Distance \\
\bottomrule
\end{tabular}
\vspace{-1.0em}
\end{table}

\subsection{Computational Cost and Efficiency Analysis}

\textbf{CE Calculation Cost.} Table~\ref{tab:computational_cost} shows CE computation on 1,000 OCR output pairs ($\sim$1K chars). Edit Distance (CPU) is fastest at 0.0002s, requiring no GPU. All GPU methods achieve $<$0.1s except bge-m3 (CPU).

\noindent\textbf{VLM Ensemble Efficiency.} CE ensembles of small VLMs outperform single large models at lower cost. E.g., 3×7B ensemble (3.5× speedup, 20× token/s improvement, 8× less memory) outperforms single 70B+, while using one GPU vs. 4×80GB (detailed VLM cost in Appendix F).

\begin{table}[htbp]
\vspace{-0.5em}
  \centering
  \caption{\textbf{Computational cost of CE on 1,000 long-text pairs (average length $\sim$1K chars).}}
  \vspace{-0.8em}
  \label{tab:computational_cost}
  \setlength{\tabcolsep}{2.5pt}
  \begin{tabular}{lc|ccc|c}
  \toprule
  \multirow{2}{*}{\textbf{CE Method}} & \multirow{2}{*}{\textbf{Device}} & \multicolumn{3}{c|}{\textbf{Time Cost (s)}} & \textbf{Memory} \\
  \cmidrule(lr){3-5}
   & & \textbf{Avg} & \textbf{Med} & \textbf{Total} & \textbf{(GiB)} \\
  \midrule
  Edit Distance & CPU &
    \textbf{2.0e-4} &
    \textbf{0} &
    \textbf{1.63e-1} &
    - \\

  bge-m3 & GPU &
    3.48e-2 &
    3.11e-2 &
    3.48e1 &
    4.01 \\

  bge-m3 & CPU &
    2.65 &
    2.09 &
    2.65e3 &
    - \\
  \bottomrule
  \end{tabular}
  \vspace{-1.0em}
\end{table}
\section{Conclusion}

We propose Consensus Entropy (CE), a training-free metric that measures agreement among multiple VLMs to estimate OCR uncertainty. CE is paired with a simple ensemble-and-route procedure: when CE is low, trust the majority; when high, route the sample to a stronger model or flag for review. Experiments show that this pair outperforms single VLM and VLM-as-Judge on verification tasks while using fewer resources. 
Future work could further extend CE to other vision-language tasks \cite{wang2024latent, davoudi2025collectivereasoningllmsframework}  and incorporate divergence patterns into training methodologies~\cite{chen2025harnessingmultiplelargelanguage}. 
We hope CE and its ensemble serve as a minimal yet strong baseline for OCR uncertainty estimation.
\section*{Acknowledgments}

This work was supported in part by the Joint Funds of the National Natural Science Foundation of China (Grant No. U21B2020).

We thank Xinyue Huang, and all coauthors for the collaborative discussions and contributions. We also gratefully acknowledge Fudan University, Shanghai Jiao Tong University, and the Shanghai Innovation Institute for their support and resources.

{
    \small
    \bibliographystyle{ieeenat_fullname}
    \bibliography{main}
}

\clearpage
\def\CEAppendixIncluded{1}

\pagestyle{plain}  


\appendix
\onecolumn

\section*{\centering \LARGE{Appendix}}
We provide more details of the proposed method and additional experimental results to help better understand our paper. The appendix is organized to present comprehensive information on our experimental setup, prompting strategies, additional results, and limitations of the current approach.

\DoToC

\section{Experiment Setting}

\textbf{Implementation.} We use inference-only evaluation with diverse VLMs following \textit{vlmevalkit} \cite{vlmevalkit} defaults. Most models run on single NVIDIA H800 (80GB) GPU; largest models (e.g., Qwen2.5-VL-72B) use 2-4 GPUs. Our work involves \textbf{only inference, eliminating training and minimizing computational cost.}
Edit distance uses Levenshtein; cosine uses bge-m3~\cite{bge-m3}, applied only to OCRBench-V2.
All experiments apply temperature 0.0 (single-run) except Self-Consistency (T=0.7, 3 runs).

The experimental environment is configured with Ubuntu 22.04.2 LTS running on Linux kernel 5.10.134-16.103.al8.x86\_64 with x86\_64 architecture. PyTorch 2.6.0 is compiled with CUDA 12.4 support (+cu124), indicating compatibility with the installed CUDA toolkit. 

\subsection{Datasets for Evaluation.}

Our research utilizes three datasets: OCRBench~\cite{OCRBench}, OCRBench-V2~\cite{OCRBench}, and CCOCR~\cite{yang2024ccocr}, with the majority of our experiments conducted on OCRBench.

To align our evaluation with human preferences in real-world contexts, we also curated a unique dataset consisting of 1,000 PDF pages randomly selected from actual scenarios. These documents were processed using the state-of-the-art Qwen2.5-VL-72B model to perform OCR tasks. The resultant texts were manually compared to the original PDF pages, and each was assigned a quality score ranging from 1.0 (Perfect Match) to 0.0 (Mostly Incorrect), with varying degrees of text matching accuracy. This diverse dataset collection allows for a comprehensive assessment of OCR performance across various domains, languages, and formatting complexities, providing insights into both machine efficiency and human-centric accuracy. Detailed scoring criteria and methodology are available in Appendix C. Data cases are shown in Appendix G. Our annotated dataset is publicly available at: \url{https://huggingface.co/datasets/Aslan-mingye/OCR-Quality}.

\subsection{Baselines}
We establish two representative baseline approaches. These methods are widely adopted in current applications of VLMs, and serve as comparative references across varying levels of complexity and supervision.

\textit{VLM-as-Judge. }This baseline follows the mainstream paradigm of using large language models as evaluators to score candidate outputs. Specifically, we employ GPT4o\footnote{The specific version  of GPT4o is gpt-40-2024-11-20.}, Qwen2-VL-72B, and Qwen2-VL-7B as evaluation models, each guided by standardized prompts to assess OCR output quality. The evaluation criteria emphasize semantic correctness, visual-text alignment, and structural fidelity. 

\textit{Single Model Output. }To better highlight the performance gains introduced by Consensus Entropy, we adopt a direct single-model output strategy as a baseline. In this setting, each image is processed by a single VLM (e.g., GPT4o or Qwen2.5-VL-72B), and its raw prediction is used as the final result without post-processing. This configuration reflects common practice in many real-world OCR applications and provides a conservative lower-bound reference for evaluating the effectiveness of our framework.

\textit{Self Consitency (SC).}~\cite{wang2022self}selecting by majority voting among multiple samples from the same VLM.
We set the temperature at 0.7 to make the models inference for 3 times and choose the best predictions.

\section{Additional Experimental Results}
This section presents additional detailed analyses of our Consensus Entropy framework to complement the results discussed in the main paper. The extended results provide deeper insights into the behavior and effectiveness of different entropy calculation methods, the impact of distribution characteristics, and practical implications for deployment.

\begin{figure}[h]
  \centering
  \includegraphics[width=\linewidth]{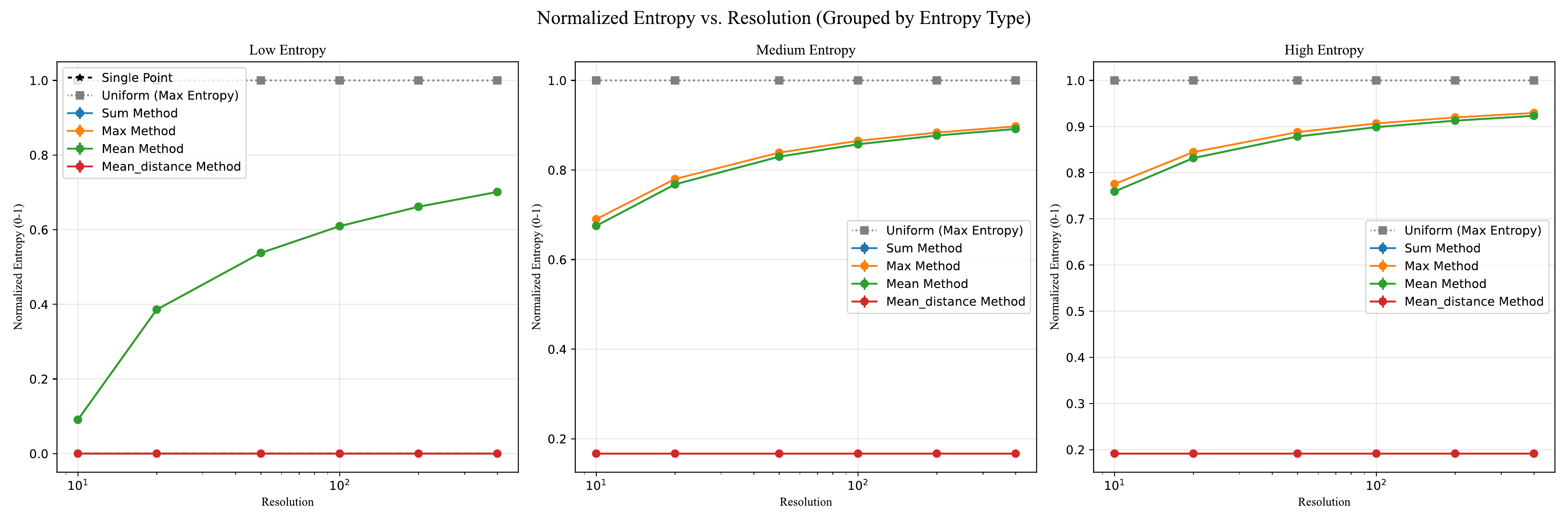}
  \caption{Normalized entropy comparison across different distribution types using the same aggregation methods. While Mean Distance shows the most distinctive pattern, all methods maintain similar relative positioning across distribution types, confirming the robustness of the entropy-based approach regardless of the specific computation method chosen.}
  \label{fig:norm_entropy_type}
\end{figure}

Figure~\ref{fig:norm_entropy_type} extends our analysis of different aggregation methods by comparing their behavior across identical distribution types. The results demonstrate that while Mean Distance exhibits the most distinctive entropy pattern, all methods maintain consistent relative entropy values across distribution types. This consistency further validates our entropy-based approach by showing that the qualitative rankings of uncertainty remain stable regardless of the specific aggregation method employed.

\subsection{Comparative Analysis of OCR Methods}
To provide a comprehensive evaluation of different OCR approaches, we conducted a detailed comparison of three main strategies: Self-Consistency (SC@3), Routing-based ensemble, and Single model performance. Table~\ref{tab:ocr_method_comparison} presents the results of this analysis, highlighting the relative performance improvements of each method.

\begin{table}[h]
  \caption{Performance comparison of different OCR methods based on cumulative scores. SC@3 represents Self-Consistency with 3 samples, Routing refers to the ensemble with rephrasing, and Single indicates individual model performance. \colorbox{tabcolor5}{Blue highlights} indicate the best performance in each category, while \colorbox{tabcolor3}{green highlights} mark the second best.}
  \centering
  \begin{tabular}{lcc}
  \toprule
  \textbf{Method} & \textbf{Score} & \textbf{Improvement (\%)} \\
  \midrule
  SC@3 (Average) & 828& -2.8\\
  SC@3 (Best: Qwen2-VL-72B) & 875& 2.7\\
  Routing (Average) & \sethlcolor{tabcolor3}\hl{\textbf{907}}& \sethlcolor{tabcolor3}\hl{\textbf{6.5}}\\
  Routing (Best: Qwen2.5-VL-72B) & \sethlcolor{tabcolor5}\hl{\textbf{922}}& \sethlcolor{tabcolor5}\hl{\textbf{8.2}}\\
  Single (Average) & 852& -\\
  Single (Best: Qwen2-VL-72B) & 888& -\\
  \bottomrule
  \end{tabular}
  \label{tab:ocr_method_comparison}
\end{table}

The results demonstrate several key findings: First, the Routing-based ensemble approach consistently outperforms both SC@3 and Single model methods, with the best routing configuration achieving a score of 922, representing an 8.2\% improvement over the best single model performance. Second, while SC@3 shows potential for improvement (2.7\% over baseline when using Qwen2-VL-72B), its average performance actually decreases by 2.8\%, indicating high variance in its effectiveness. Third, the gap between average and best performance is smallest for the Routing method (15 points) compared to SC@3 (47 points) and Single models (36 points), suggesting more consistent and reliable performance across different configurations.

\begin{figure}[h]
  \centering
  \includegraphics[width=\linewidth]{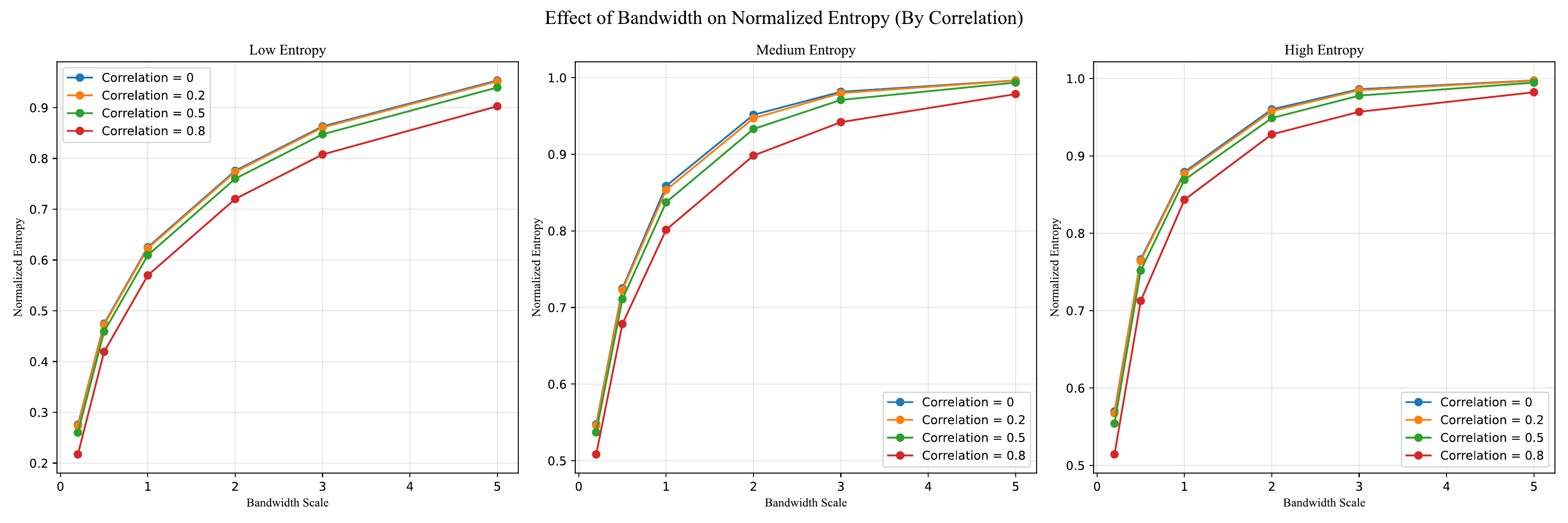}
  \caption{Effects of kernel bandwidth and distribution correlation on entropy estimates. Higher correlation values consistently result in lower entropy curves, highlighting how the spatial relationship between predictions influences the consensus measurement.}
  \label{fig:bandwidth_corr}
\end{figure}

Figure~\ref{fig:bandwidth_corr} investigates how kernel bandwidth settings and the correlation between distributions affect entropy estimates. A key finding is that higher correlation values consistently produce lower entropy curves, indicating that tightly clustered predictions (as would be expected for correct OCR outputs) naturally result in lower entropy measurements. This property is precisely what enables our CE metric to effectively distinguish between high-confidence and low-confidence predictions.

\begin{figure}[h]
  \centering
  \includegraphics[width=\linewidth]{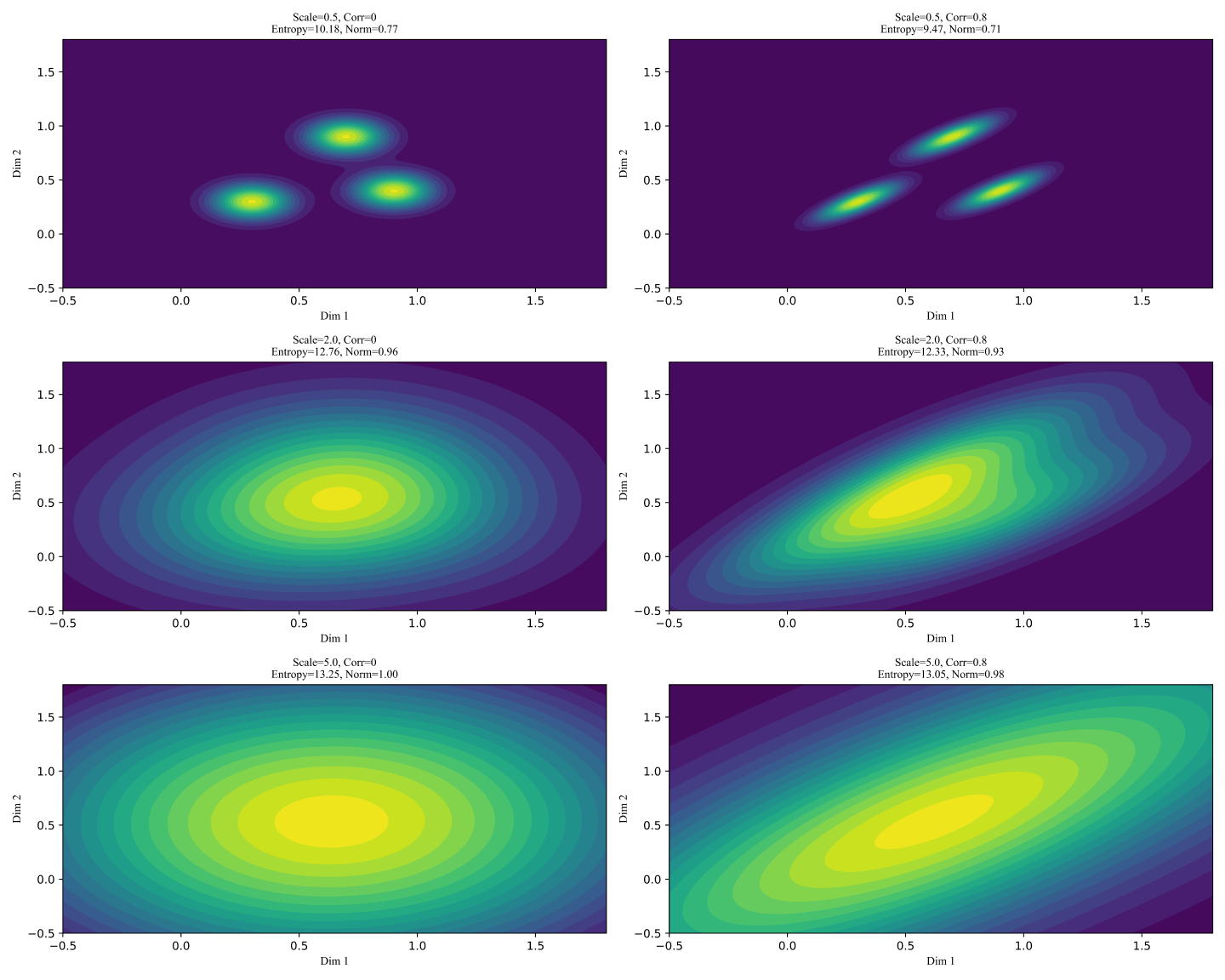}
  \caption{Visualization of entropy patterns across various scale and correlation settings. These examples illustrate how our entropy calculation responds to different distribution characteristics, with tighter, more correlated clusters resulting in lower entropy values regardless of scale.}
  \label{fig:dist_samples}
\end{figure}

We also conducted experiments on OCRBench-V2 regarding the relationship between CE thresholds and benchmark scores, as shown in Table~\ref{tab:v2_scores}.

\subsection{CE Thresholds Analysis}

Table~\ref{tab:v2_scores} reports the average scores of three models on OCRBench-V2 under different CE thresholds, showing that CE remains effective for verifying OCR and VQA-style tasks beyond the main OCRBench setting.
Here, CE is used as a routing signal: lower thresholds keep only low-entropy (high-confidence) predictions, while higher thresholds allow more samples to pass.

To make the routing behaviour more explicit, we further summarize the accuracy–computation trade-off in terms of \emph{accuracy improvement vs.\ rephrase ratio}.
Table~\ref{tab:threshold_sensitivity_appendix} (numbers reproduced from the main paper and rebuttal) shows, for representative models on OCRBench-V2, the gain in accuracy ($\Delta \times 10^2$) and the percentage of samples routed for rephrasing at different thresholds.
Smaller $\theta$ values route more samples and yield higher accuracy but incur greater computation, while larger $\theta$ values are more efficient but less aggressive.
For CE-OCR experiments on OCRBench, we use a single global threshold $\theta = 0.5$ (Section 4.3); for OCRBench-V2 routing, the sweep in Table~\ref{tab:threshold_sensitivity_appendix} indicates that values around $\theta \approx 0.5$ offer a good balance between accuracy gains and routing cost.

\begin{table}[h]
\centering
\caption{Threshold sensitivity analysis on OCRBench-V2. Each cell shows accuracy improvement ($\Delta \times 10^2$) / rephrase percentage (\%). Lower $\theta$ routes more samples for rephrasing, improving accuracy at the cost of additional computation.}
\label{tab:threshold_sensitivity_appendix}
\small
\setlength{\tabcolsep}{3pt}
\begin{tabular}{l|c|c|c|c|c}
\toprule
\textbf{Model} & $\theta=0.95$ & $\theta=0.9$ & $\theta=0.8$ & $\theta=0.6$ & $\theta=0.2$ \\
\midrule
Gemini Pro & +1.6 / 11.5 & +3.1 / 22.3 & +4.3 / 43.5 & +5.4 / 66.3 & +6.1 / 91.5 \\
GPT-4o & +3.8 / 15.0 & +5.1 / 24.9 & +6.8 / 48.7 & +7.7 / 71.3 & +8.8 / 91.2 \\
\bottomrule
\end{tabular}
\end{table}

\subsection{Model Calibration Analysis}
\label{sec:appendix_calibration}

\textbf{Note:} ECE and Brier Score metrics are designed for classification probability calibration and are mismatched for our generation-based approach. \textbf{This analysis is provided for reference only and should not be used as the primary evaluation criterion for CE, which is a post-hoc uncertainty metric derived from output agreement rather than probability prediction.}

For completeness, we evaluated CE using Expected Calibration Error (ECE), Brier Score, and AUC metrics against VLM-as-Judge methods using Qwen2.5VL-72B with different reference model combinations (diverse: InternVL+GPT4o vs related: QwenVL series). Despite ECE and Brier scores being primarily designed for classification probabilities, CE achieves favorable metrics (best ECE: 0.0842, Brier: 0.1169, AUC: 0.9226), though these values should not be overinterpreted given the methodological mismatch.

\begin{table}[h]
\centering
\caption{Calibration metrics for CE and VLM-as-Judge on Qwen2.5VL-72B predictions. CE demonstrates better calibration despite being designed for generation tasks rather than classification.}
\label{tab:calibration_metrics}
\begin{tabular}{lccc}
\toprule
\textbf{Method} & \textbf{ECE}↓ & \textbf{Brier}↓ & \textbf{AUC}↑ \\
\midrule
Judge-Related & 0.1090 & 0.1404 & 0.8041 \\
Judge-Diverse & 0.0952 & 0.1392 & 0.7727 \\
CE-Related & 0.0938 & 0.1362 & 0.9181 \\
CE-Diverse & \textbf{0.0842} & \textbf{0.1169} & \textbf{0.9226} \\
\bottomrule
\end{tabular}
\end{table}

\subsection{Identical Model Sampling Analysis}
\label{sec:appendix_identical_sampling}

To evaluate CE's robustness with minimal model diversity, we conducted experiments using identical models with stochastic sampling (temperature=0.7, 3 runs). Table~\ref{tab:identical_model_sampling} shows that even with the same model architecture and weights, CE-Ensemble consistently improves performance by leveraging sampling variance. This demonstrates that CE captures task-level uncertainty effectively, even when architectural diversity is absent.

\begin{table}[h]
\centering
\caption{CE-Ensemble performance with identical models using stochastic sampling. Individual run scores shown in parentheses. CE consistently selects better outputs across sampling variance.}
\label{tab:identical_model_sampling}
\small
\setlength{\tabcolsep}{2pt}
\begin{tabular}{lcccccc}
\toprule
\textbf{Model} & \textbf{Text Rec.} & \textbf{Scene VQA} & \textbf{Doc VQA} & \textbf{KIE} & \textbf{Formula} & \textbf{ALL} \\
\midrule
Qwen2VL-7B & 272 & 173 & 155 & 185 & 67 & 852 \\
 & (266,266,269) & (171,168,170) & (156,151,149) & (182,181,181) & (59,61,63) & (834,827,832) \\
\midrule
Qwen2VL-72B & 264 & 181 & 173 & 186 & 71 & 875 \\
 & (265,263,261) & (179,182,177) & (166,170,173) & (183,183,183) & (61,64,71) & (854,865,865) \\
\midrule
Qwen2.5VL-72B & 264 & 176 & 175 & 184 & 71 & 870 \\
 & (256,260,264) & (172,174,174) & (171,169,173) & (184,182,186) & (71,69,67) & (854,854,864) \\
\bottomrule
\end{tabular}
\end{table}

\subsection{GPT-4o Output Verbosity Analysis}
\label{sec:appendix_gpt4o_verbosity}

During threshold analysis (Figure 4 in main paper), GPT-4o exhibited non-monotonic accuracy patterns. Investigation revealed that GPT-4o generates verbose outputs for short-answer questions in OCRBench. For example, when other models answer ``11.90'', GPT-4o responds: ``The total amount of this receipt is **RM 11.90**, as stated under `total incl. GST @6\%'''. This verbosity artificially inflates CE values. Table~\ref{tab:gpt4o_cleaning} shows that after cleaning outputs via regex and LLM-based filtering, GPT-4o's accuracy exhibits expected monotonicity with CE thresholds.

\begin{table}[h]
\centering
\caption{GPT-4o performance under different CE thresholds before and after output cleaning. Cleaned outputs restore monotonic relationship between CE and accuracy.}
\label{tab:gpt4o_cleaning}
\begin{tabular}{lcc}
\toprule
\textbf{CE Threshold} & \textbf{Original} & \textbf{Cleaned} \\
\midrule
$\leq$0.7 & 0.781 & 0.831 \\
$\leq$0.4 & 0.833 & 0.884 \\
$\leq$0.2 & 0.833 & 0.928 \\
$\leq$0.1 & 0.844 & 0.930 \\
\bottomrule
\end{tabular}
\end{table}

\subsection{Distance Metric Comparison}
\label{sec:appendix_distance_metrics}

CE supports multiple distance metrics depending on task requirements. Table~\ref{tab:distance_metric_comparison} compares Edit Distance (character-level) and Cosine Distance (semantic-level) across different task types. Edit Distance excels in pure OCR tasks requiring character-level precision, while Cosine Distance (using bge-m3) performs better on complex VQA tasks with semantic variations. Both maintain the same CE framework—only the pairwise similarity computation differs.

\begin{table}[h]
\centering
\caption{Comparison of distance metrics for CE computation. The choice of metric depends on task characteristics: character-level precision vs semantic understanding.}
\label{tab:distance_metric_comparison}
\small
\setlength{\tabcolsep}{3pt}
\begin{tabular}{lllll}
\toprule
\textbf{Method} & \textbf{Granularity} & \textbf{Best For} & \textbf{Cost} & \textbf{Used In} \\
\midrule
Edit Distance & Character & OCR, Math, Code, & Very low & Main OCRBench \\
 &  & Simple VQA & (CPU) & experiments; Table 1 \\
\midrule
Cosine & Semantic & Complex VQA, & Low (small & Appendix \\
Distance &  & Diverse questions & model) & OCRBench-V2; Table 4 \\
\bottomrule
\end{tabular}
\end{table}

\begin{longtable}{|l|c|c|c|c|c|}
\caption{Performance of 3 models on OCRBench-V2 under Different CE Thresholds. }
\small
\label{tab:v2_scores}\\
\hline
\textbf{Model} & \textbf{CE Threshold} & \textbf{Total QAs} & \textbf{EN Overall} & \textbf{CN Overall} & \textbf{ALL Overall} \\
\hline
\endfirsthead

\multicolumn{6}{c}%
{{\bfseries \tablename\ \thetable{} -- continued from previous page}} \\
\hline
\textbf{Model} & \textbf{Threshold} & \textbf{Total Questions} & \textbf{EN Overall} & \textbf{CN Overall} & \textbf{ALL Overall} \\
\hline
\endhead

\hline
\multicolumn{6}{|r|}{{Continued on next page}} \\
\hline
\endfoot

\hline
\endlastfoot

\multirow{11}{*}{\textbf{Gemini-PRO}} 
& 1.0 & 10000 & 0.519 & 0.431 & 0.520 \\
& 0.9 & 8816 & 0.549 & 0.465 & 0.546 \\
& 0.8 & 7252 & 0.580 & 0.535 & 0.582 \\
& 0.7 & 6009 & 0.614 & 0.560 & 0.610 \\
& 0.6 & 5221 & 0.643 & 0.588 & 0.633 \\
& 0.5 & 4499 & 0.640 & 0.596 & 0.652 \\
& 0.4 & 3276 & 0.702 & 0.650 & 0.668 \\
& 0.3 & 2323 & 0.772 & 0.698 & 0.701 \\
& 0.2 & 1664 & 0.785 & 0.648 & 0.743 \\
& 0.1 & 1041 & 0.828 & 0.690 & 0.790 \\
\hline

\multirow{11}{*}{\textbf{GPT4o}} 
& 1.0 & 10000 & 0.465 & 0.322 & 0.473 \\
& 0.9 & 8357 & 0.521 & 0.394 & 0.541 \\
& 0.8 & 6571 & 0.559 & 0.463 & 0.584 \\
& 0.7 & 5341 & 0.585 & 0.529 & 0.604 \\
& 0.6 & 4566 & 0.595 & 0.581 & 0.615 \\
& 0.5 & 3903 & 0.614 & 0.625 & 0.639 \\
& 0.4 & 3031 & 0.652 & 0.647 & 0.652 \\
& 0.3 & 2423 & 0.665 & 0.708 & 0.682 \\
& 0.2 & 1665 & 0.769 & 0.663 & 0.736 \\
& 0.1 & 1192 & 0.823 & 0.740 & 0.796 \\
\hline

\multirow{11}{*}{\textbf{Intern2.5VL-26B}} 
& 1.0 & 10000 & 0.494 & 0.442 & 0.532 \\
& 0.9 & 7160 & 0.558 & 0.511 & 0.580 \\
& 0.8 & 5948 & 0.581 & 0.553 & 0.593 \\
& 0.7 & 5198 & 0.614 & 0.551 & 0.620 \\
& 0.6 & 4629 & 0.626 & 0.623 & 0.642 \\
& 0.5 & 3976 & 0.634 & 0.646 & 0.664 \\
& 0.4 & 2850 & 0.673 & 0.630 & 0.697 \\
& 0.3 & 2056 & 0.677 & 0.658 & 0.744 \\
& 0.2 & 1638 & 0.776 & 0.696 & 0.772 \\
& 0.1 & 1118 & 0.829 & 0.742 & 0.799 \\
\hline
\end{longtable}

\section{RealData Source}
The datasets used in our OCR evaluation experiments were carefully curated to reflect diverse real-world document scenarios. Below we present the detailed composition of these datasets, encompassing various languages, document types, and content domains.

\begin{table}[h]
\centering
\caption{Sources of the human-labeled OCR Dataset}
\label{tab:data_sources}
\begin{tabularx}{\textwidth}{lXXXX}
\toprule
& \textbf{ebook} & \textbf{paper} & \textbf{textbook} & \textbf{other} \\
\midrule
\textbf{ZH} & zhishilei, zhongwenzaixian, gift, thomas & - & by, kps, kmath, zhonggaokao, gaojiaoshe, gaodengjiaoyu, k12 edu platform, jiaoan, zju icicles & - \\
\textbf{EN} & theeye, physicsandmathstutor, planetebook & escholarship, biorxiv, springer, sagepub, scholarworks, psyarxiv, chemrxiv, iopscience, royalsocietypublishing, criso & kps, bookboon, california 14sets, scholarworks & dev books repository \\
\textbf{ML} & renhang, banshujiang & - & openstax, math & - \\
\textbf{Other} & - & - & - & coursehero, studypool \\
\bottomrule
\end{tabularx}
\end{table}

The dataset composition encompasses a wide range of content types and sources, ensuring comprehensive evaluation of OCR capabilities. Chinese language materials include electronic books from major repositories as well as educational content across various academic levels. English content spans scholarly publications from multiple disciplines, educational materials, and literature. Multilingual resources incorporate specialized content with mixed language elements, while additional sources provide diverse formatting challenges. This heterogeneous collection enables robust assessment of OCR performance across domains, languages, and formatting complexities representative of real-world document processing requirements.

For human evaluation of OCR quality, we developed a standardized 4-level scoring system:
\begin{enumerate}
\item \textbf{Perfect Match (0.9-1.0)}: The prediction matches the image text exactly with no errors.
\item \textbf{Minor Errors (0.7-0.8)}: Very close to the image text with only small mistakes that do not affect understanding.
\item \textbf{Partially Correct (0.4-0.6)}: Contains noticeable errors or captures only part of the text.
\item \textbf{Mostly Incorrect (0.0-0.3)}: Largely incorrect or unrelated to the text in the image.
\end{enumerate}
This scoring system was used consistently across all human evaluations in our experiments to ensure reliable assessment of OCR quality.

\section{Annotation Interface for Human Evaluation}
\begin{figure}[htpt]
  \centering
  \includegraphics[width=\linewidth]{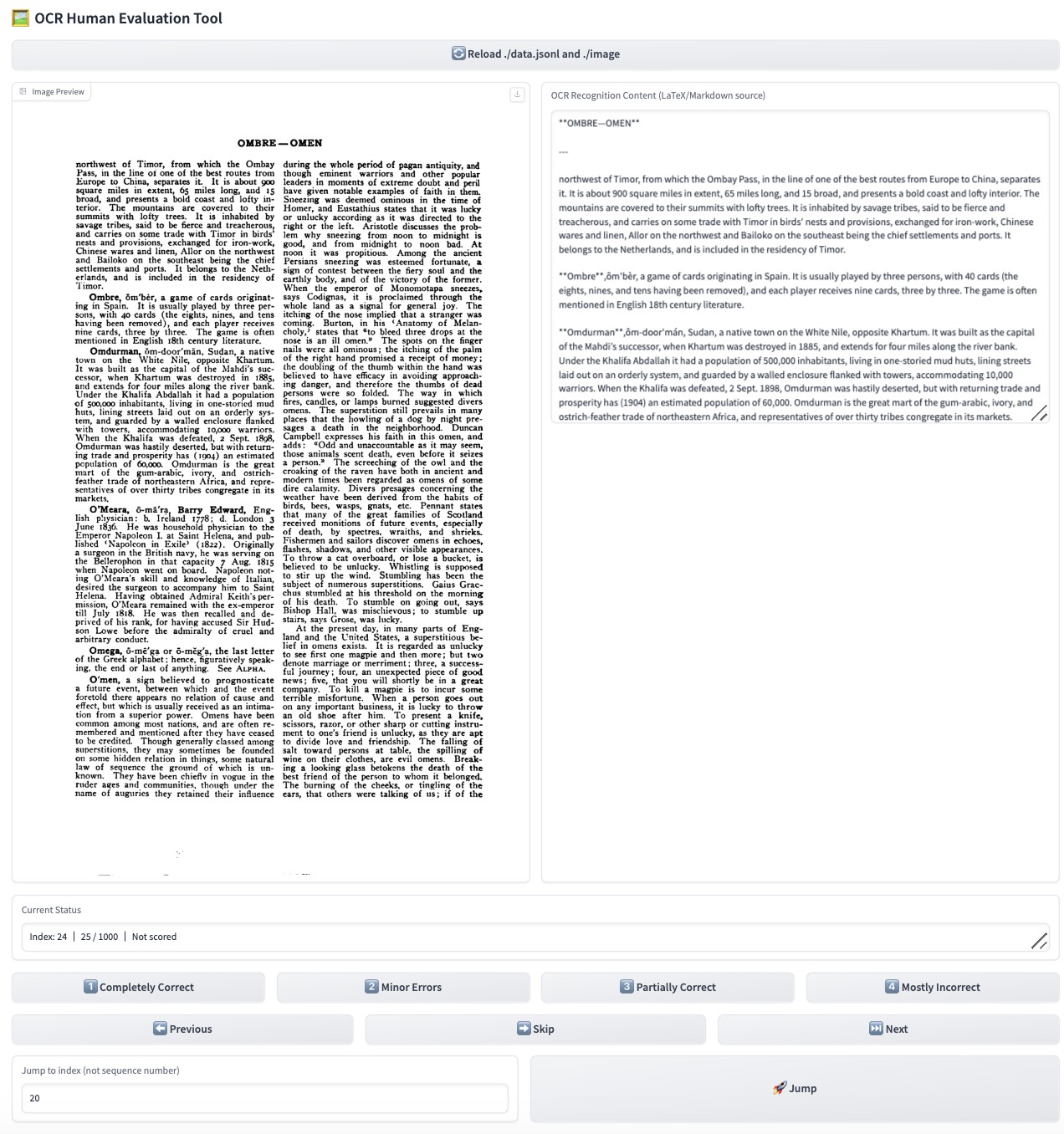}
  \caption{OCR Human Evaluation Interface: This annotation tool enables efficient assessment of OCR quality across multiple models. The interface displays the source image alongside OCR results, allowing annotators to assign quality scores based on a standardized 4-level rating system (from completely correct to mostly incorrect). The tool tracks progress, enables navigation between samples, and supports targeted image selection, facilitating comprehensive human evaluation of OCR performance on real-world data.}
  \label{fig:annotation_ui}
\end{figure}

To ensure the quality of our real-world dataset and provide reliable human judgments for OCR performance evaluation, we developed a dedicated annotation interface. Figure~\ref{fig:annotation_ui} presents our custom-built tool designed specifically for OCR evaluation tasks.

The annotation interface is implemented using Gradio, a Python library for creating web-based interfaces. This tool facilitates efficient human evaluation of OCR results by displaying the original image alongside the extracted text. Annotators can assign one of four quality ratings to each OCR output: (1) Completely Correct, (2) Minor Errors, (3) Partially Correct, or (4) Mostly Incorrect. This standardized rating system ensures consistency across evaluations and provides fine-grained assessment of OCR quality.

The interface includes several features to enhance annotation efficiency: a status display showing progress through the dataset, navigation controls to move between samples, the ability to skip difficult cases, and a direct jump function to specific image indices. All annotations are automatically saved to the original data file, creating a persistent record of human judgments. This annotation tool played a critical role in developing our ground truth evaluations and validating the effectiveness of the Consensus Entropy framework against human quality assessments.
\newpage
\section{Algorithm Details}
This section provides detailed pseudocode for the key algorithms used in our Consensus Entropy framework.

\begin{algorithm}
\caption{Consensus Entropy for Multi-Model OCR Evaluation}
\label{alg:consensus_entropy}
\begin{algorithmic}
\REQUIRE $\{O_1, O_2, \ldots, O_n\}$ - outputs from $n$ OCR models for image $I$
\REQUIRE $\sigma$ - base kernel bandwidth parameter
\REQUIRE $\alpha$ - adaptive bandwidth scaling factor
\REQUIRE $N$ - grid resolution for probability density estimation
\STATE // Compute pairwise entropies and weights
\FOR{$i = 1$ to $n$}
    \FOR{$j = 1$ to $n$, $j \neq i$}
        \STATE $\mathcal{E}_{ij} \leftarrow \mbox{ComputePairwiseEntropy}(O_i, O_j)$
    \ENDFOR
    \STATE $\overline{\mathcal{E}}_{i} \leftarrow \frac{1}{n-1} \sum_{j \neq i} \mathcal{E}_{ij}$ \COMMENT{Average entropy distance}
    \STATE $w_i \leftarrow \frac{1/\overline{\mathcal{E}}_{i}}{\sum_{k=1}^{n} 1/\overline{\mathcal{E}}_{k}}$ \COMMENT{Normalize weights}
    \STATE $\Sigma_i \leftarrow \sigma \cdot (1 + \alpha \cdot \overline{\mathcal{E}}_{i}) \cdot \mathbf{I}$ \COMMENT{Adaptive covariance}
\ENDFOR
\STATE // Project outputs to semantic space and estimate distribution on N×N grid
\STATE $\{\mathbf{v}_1, \mathbf{v}_2, \ldots, \mathbf{v}_n\} \leftarrow \mbox{ProjectToSemanticSpace}(\{O_1, O_2, \ldots, O_n\})$
\STATE $P \leftarrow \mbox{EstimateDistributionOnGrid}(\{\mathbf{v}_i, w_i, \Sigma_i\}_{i=1}^n, N)$
\STATE // Calculate entropy from $N \times N$ grid-based distribution
\STATE $\delta \leftarrow -\sum_{x=1}^{N} \sum_{y=1}^{N} P[x,y] \cdot \log P[x,y]$ \COMMENT{Discrete entropy calculation on $N \times N$ grid}
\RETURN $\delta$ \COMMENT{Return Consensus Entropy}
\end{algorithmic}
\end{algorithm}

\begin{algorithm}
\caption{CE-OCR: Entropy-guided Ensemble and Routing Framework}
\label{alg:consensus_entropy_framework}
\begin{algorithmic}
\REQUIRE $I$ - input image containing text
\REQUIRE $\{M_1, M_2, \ldots, M_n\}$ - set of OCR models
\REQUIRE $M_{\text{exp}}$ - stronger vision-language model for rephrasing
\REQUIRE $\theta$ - entropy threshold for routing
\STATE $\{O_1, O_2, \ldots, O_n\} \leftarrow \{M_1(I), M_2(I), \ldots, M_n(I)\}$ \COMMENT{Generate predictions}
\STATE $\delta \leftarrow \text{ConsensusEntropy}(\{O_1, O_2, \ldots, O_n\})$ \COMMENT{Calculate entropy using Algorithm 1}
\IF{$\delta \leq \theta$}
    \STATE // Use precomputed weights from Algorithm 1
    \STATE $O_{\text{final}} \leftarrow \text{WeightedEnsemble}(\{O_1, O_2, \ldots, O_n\}, \{w_1, w_2, \ldots, w_n\})$
\ELSE
    \STATE $O_{\text{ens}} \leftarrow \text{SimpleEnsemble}(\{O_1, O_2, \ldots, O_n\})$ \COMMENT{Basic ensemble}
    \STATE $O_{\text{final}} \leftarrow M_{\text{exp}}(I, \{O_1, O_2, \ldots, O_n\}, O_{\text{ens}})$ \COMMENT{Route to stronger model for rephrasing}
\ENDIF
\RETURN $O_{\text{final}}$
\end{algorithmic}
\end{algorithm}

\subsection{CE Hyperparameters and Settings}

For reproducibility, we summarize the key hyperparameters used in all CE experiments.

\textbf{Distance Metrics.} As detailed in Table~\ref{tab:distance_metric_comparison}, CE supports two distance metrics:
(1) character-level Edit Distance for pure OCR tasks, mathematical formulas, code, and simple VQA, and
(2) semantic Cosine Distance for complex VQA and diverse questions.
In practice, all OCRBench (main paper Table 1) and human-evaluation experiments use Edit Distance on raw text strings, while OCRBench-V2 experiments employ Cosine Distance based on bge-m3 embeddings (and bge-large-zh-v1.5 in early pilot experiments), following the configurations discussed in the rebuttal.

\textbf{Kernel and Grid Parameters.} In the semantic case, we apply isotropic Gaussian kernels with covariance
$\Sigma_i = \sigma \cdot (1 + \alpha \cdot \overline{\mathcal{E}}_{i}) \cdot \mathbf{I}$ on a fixed $N \times N$ grid, as specified in Algorithm~\ref{alg:consensus_entropy}. 
The bandwidth base $\sigma$, scaling factor $\alpha$, and grid resolution $N$ are treated as global constants shared by all models on a given benchmark and tuned once on a held-out development split.
We observe that CE is empirically robust to moderate changes in these values; performance is dominated by the choice of distance metric and routing threshold rather than fine-grained kernel tuning.

\textbf{Sampling Temperature.} Unless otherwise stated, all VLMs run with temperature $0.0$ (greedy decoding).
Self-Consistency baselines and identical-model CE-Ensemble experiments use temperature $0.7$ with 3 samples, as reported in Section 4 and Table~\ref{tab:identical_model_sampling}.

\section{Computational Cost and Efficiency Analysis}

We measure both CE calculation costs and efficiency, comparing single-query verification (requiring one pairwise CE computation) against ensemble approaches.

\textit{Computational Cost of CE.} We benchmarked CE computation on 1,000 OCR output pairs (average length $\sim$1K characters). Table~\ref{tab:computational_cost} presents detailed computational cost analysis. Edit Distance (CPU) is the fastest method at 0.0002s per computation, requiring no GPU resources. All methods except bge-m3 (CPU) achieve sub-0.1s computation times. For maximum efficiency, we recommend using Edit Distance for character-level tasks and GPU-accelerated models for semantic tasks, while avoiding CPU execution of large models.

\textit{VLM Cost Comparison.}
Paper "Figure 5: Performance comparison across token lengths" show that CE based on small VLMs \textbf{outperform stronger single models and require fewer tokens, less GPU memory, and offer faster inference.} Notably, while 70B+ models demand at least 4$\times$80GB GPUs, multiple small models can run concurrently on a single GPU, highlighting the efficiency of the CE framework.

\begin{table}[h]
  \centering
  \setlength{\tabcolsep}{2.5pt}
  \begin{tabular}{l|ccc|c}
  \toprule
  \multirow{2}{*}{\textbf{CE Method}} & \multicolumn{3}{c|}{\textbf{Time Cost (s)}} & \multirow{2}{*}{\textbf{Memory (GiB)}} \\
   & \textbf{Avg} & \textbf{Med} & \textbf{Total} & \\
  \midrule
  Edit Distance (CPU) & \textbf{0.0002} & \textbf{0.0000} & \textbf{0.1626} & - \\
  bge1.5-zh (GPU) & 0.0116 & 0.0117 & 11.5716 & 1.19 \\
  bge-m3 (GPU) & 0.0348 & 0.0311 & 34.7902 & 4.01 \\
  bge-m3 (CPU) & 2.6518 & 2.0933 & 2651.7895 & - \\
  \midrule
  \end{tabular}
  \caption{
    \textbf{Computational cost of CE on 1,000 long-text pairs (average length $\sim$1K chars).} Avg/Med/Total: average,median, total time cost of one pair.
    }
  \label{tab:computational_cost}
  \end{table}
  
\begin{table}[h]
    \caption{VLM cost comparison on OCRBench. Ensembles of small models achieve better performance with lower resource requirements than single large models.}
    \centering
    \begin{tabular}{lccc}
    \toprule
    \textbf{Model} & \textbf{Time (s)} & \textbf{Token/s} & \textbf{Memory (GiB)} \\
    \midrule
    InternVL2.5-4B & 409 & 6.98 & 7.90 \\
    InternVL2.5-8B & 672 & 4.28 & 20.71 \\
    InternVL2.5-26B & 951 & 3.03 & 52.67 \\
    InternVL2.5-78B & 1772 & 1.62 & 158.53 \\
    Ovis2-4B & 680 & 27.49 & 9.09 \\
    Ovis2-8B & 611 & 30.27 & 18.26 \\
    Ovis2-34B & 1191 & 15.33 & 66.60 \\
    Qwen2.5VL-3B & 633 & 19.05 & 8.04 \\
    Qwen2.5VL-7B & 771 & 153.16 & 16.99 \\
    Qwen2.5VL-72B & 2715 & 7.52 & 140.98 \\
    \midrule
    \end{tabular}
    \label{tab:vlm_cost_comparison}
    \end{table}

\newpage
\section{OCR Evaluation Case Studies: CE vs. VLM-as-Judge Comparison}
\label{sec:ocr_case_studies}

To provide a more intuitive understanding of how Consensus Entropy (CE) compares with VLM-as-Judge methods in real-world OCR tasks, we present several representative cases that highlight their respective performance characteristics.

\noindent\textbf{Scoring Methodology.} All cases use the 4-level scoring system defined in Appendix C: Perfect Match (0.9--1.0), Minor Errors (0.7--0.8), Partially Correct (0.4--0.6), and Mostly Incorrect (0.0--0.3). For CE interpretation: low CE ($<0.3$) indicates high consensus and likely correct output; medium CE ($0.3$--$0.7$) suggests moderate disagreement; high CE ($>0.7$) signals significant divergence and probable errors. 

\begin{tcolorbox}[colback=casebg, colframe=black, boxrule=1pt, arc=3mm, left=2mm, right=2mm, top=2mm, bottom=2mm]
\begin{minipage}{\linewidth}
    \centering
    \includegraphics[width=1.0\linewidth]{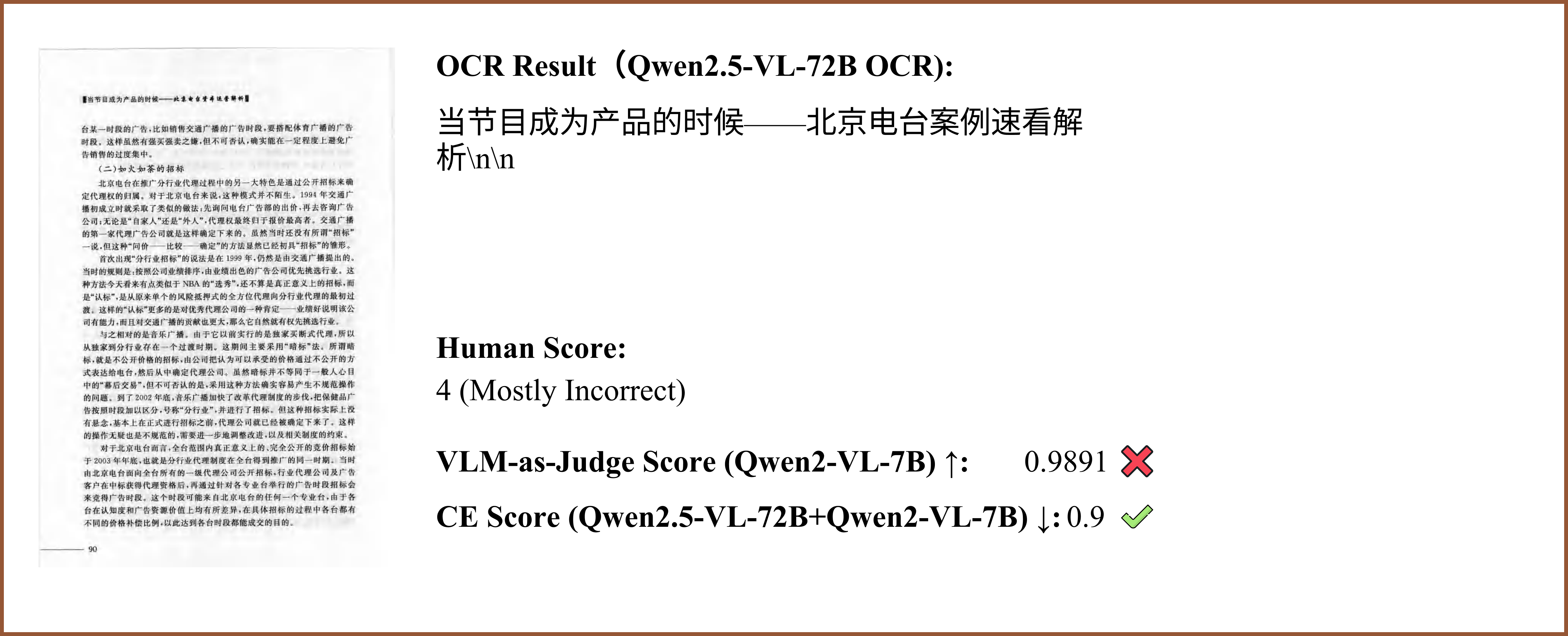}
    \captionof{figure}{Case 1: High-quality Chinese OCR sample with strong agreement between human evaluation (4/4), CE score (0.9891), and VLM-as-Judge score (0.9).}
    \label{fig:case1}
\end{minipage}
\end{tcolorbox}

\begin{tcolorbox}[colback=casebg, colframe=black, boxrule=1pt, arc=3mm, left=2mm, right=2mm, top=2mm, bottom=2mm]
\begin{minipage}{\linewidth}
    \centering
    \includegraphics[width=1.0\linewidth]{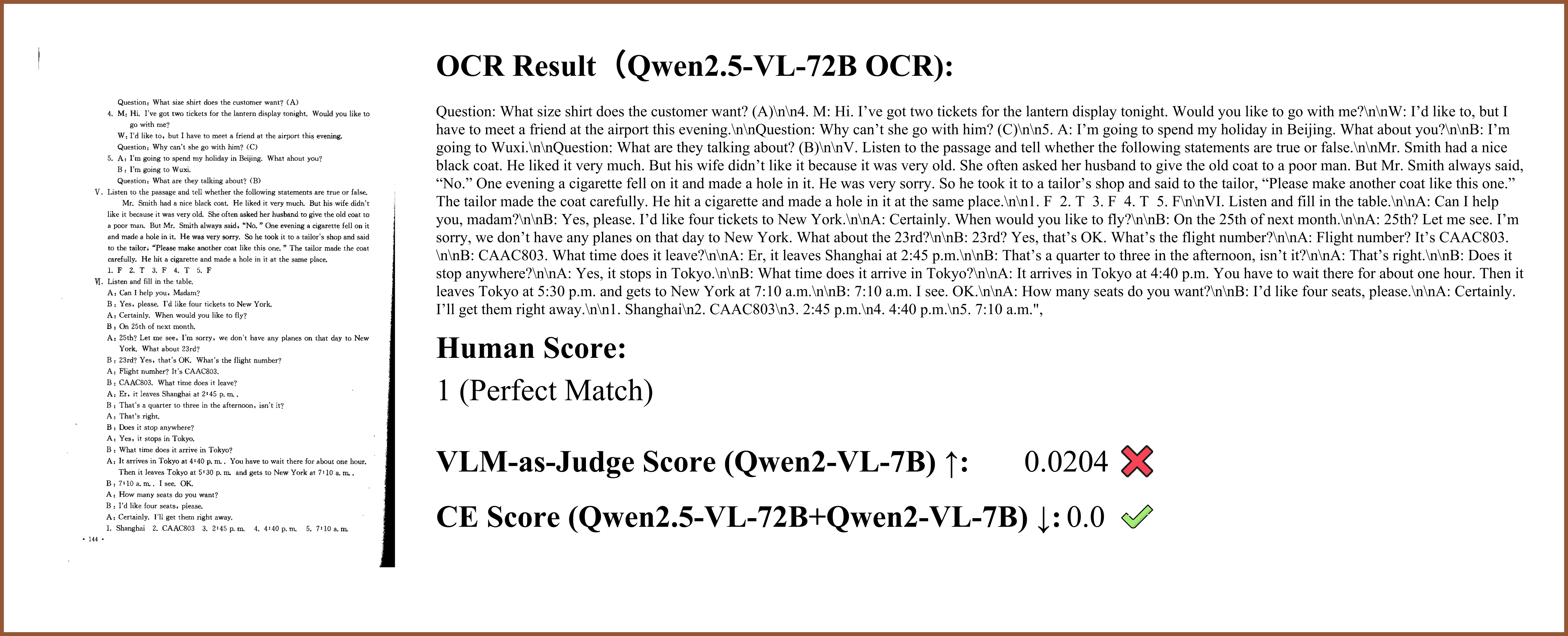}
    \captionof{figure}{Case 2: Low-quality English OCR sample with unanimous poor quality assessment across human evaluation (1/4), CE score (0.0204), and VLM-as-Judge score (0.0).}
    \label{fig:case2}
\end{minipage}
\end{tcolorbox}

\noindent\textbf{Case 2 Analysis.} Although the OCR output in Case 2 may appear superficially readable, it contains critical errors in financial and numerical figures that cause high VLM disagreement. These character-level discrepancies---easily missed by semantic-only methods---result in elevated CE values, correctly flagging the output as unreliable for downstream applications such as invoice processing or data extraction.

\begin{tcolorbox}[colback=casebg, colframe=black, boxrule=1pt, arc=3mm, left=2mm, right=2mm, top=2mm, bottom=2mm]
\begin{minipage}{\linewidth}
    \centering
    \includegraphics[width=1.0\linewidth]{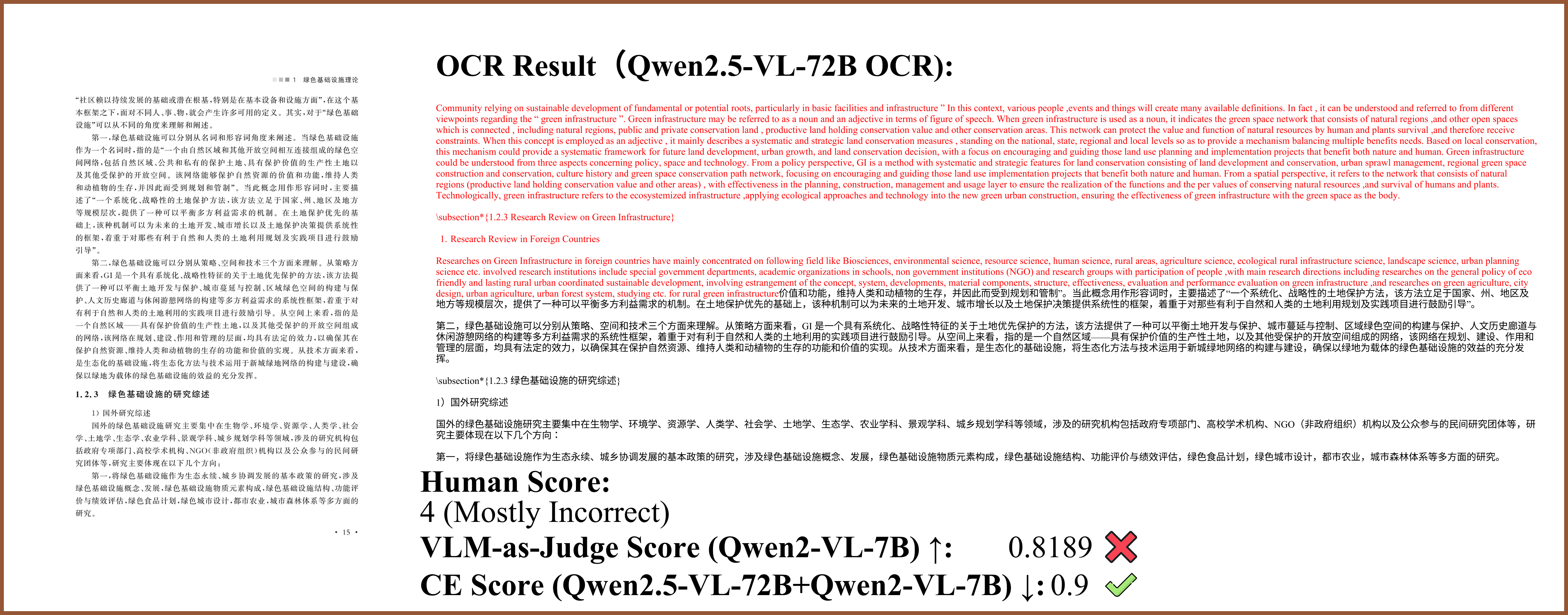}
    \captionof{figure}{Case 3: High-quality academic Chinese text correctly identified by all evaluation methods, demonstrating strong performance on complex domain-specific content.}
    \label{fig:case3}
\end{minipage}
\end{tcolorbox}

\begin{tcolorbox}[colback=casebg, colframe=black, boxrule=1pt, arc=3mm, left=2mm, right=2mm, top=2mm, bottom=2mm]
\begin{minipage}{\linewidth}
    \centering
    \includegraphics[width=1.0\linewidth]{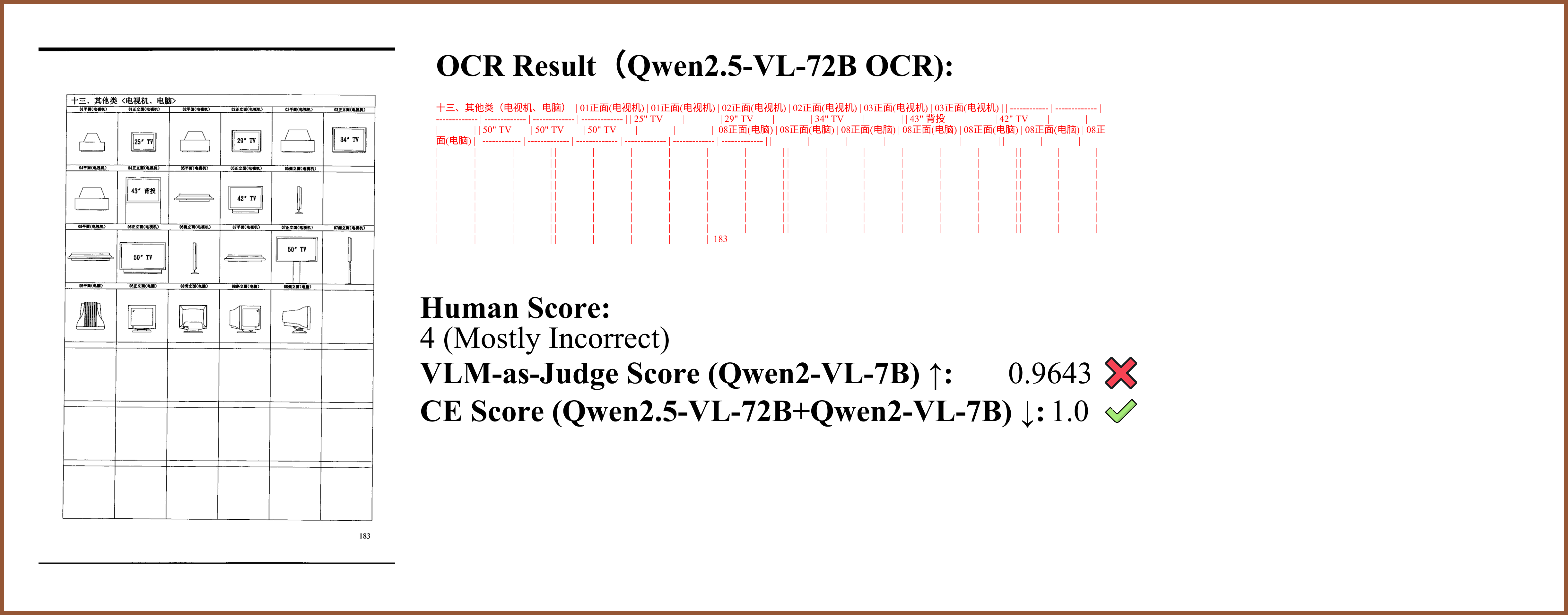}
    \captionof{figure}{Case 4: Structured tabular content with perfect human score (4/4) and near-perfect automated evaluation scores from both CE (0.9643) and VLM-as-Judge (1.0).}
    \label{fig:case4}
\end{minipage}
\end{tcolorbox}

\begin{tcolorbox}[colback=casebg, colframe=black, boxrule=1pt, arc=3mm, left=2mm, right=2mm, top=2mm, bottom=2mm]
\begin{minipage}{\linewidth}
    \centering
    \includegraphics[width=1.0\linewidth]{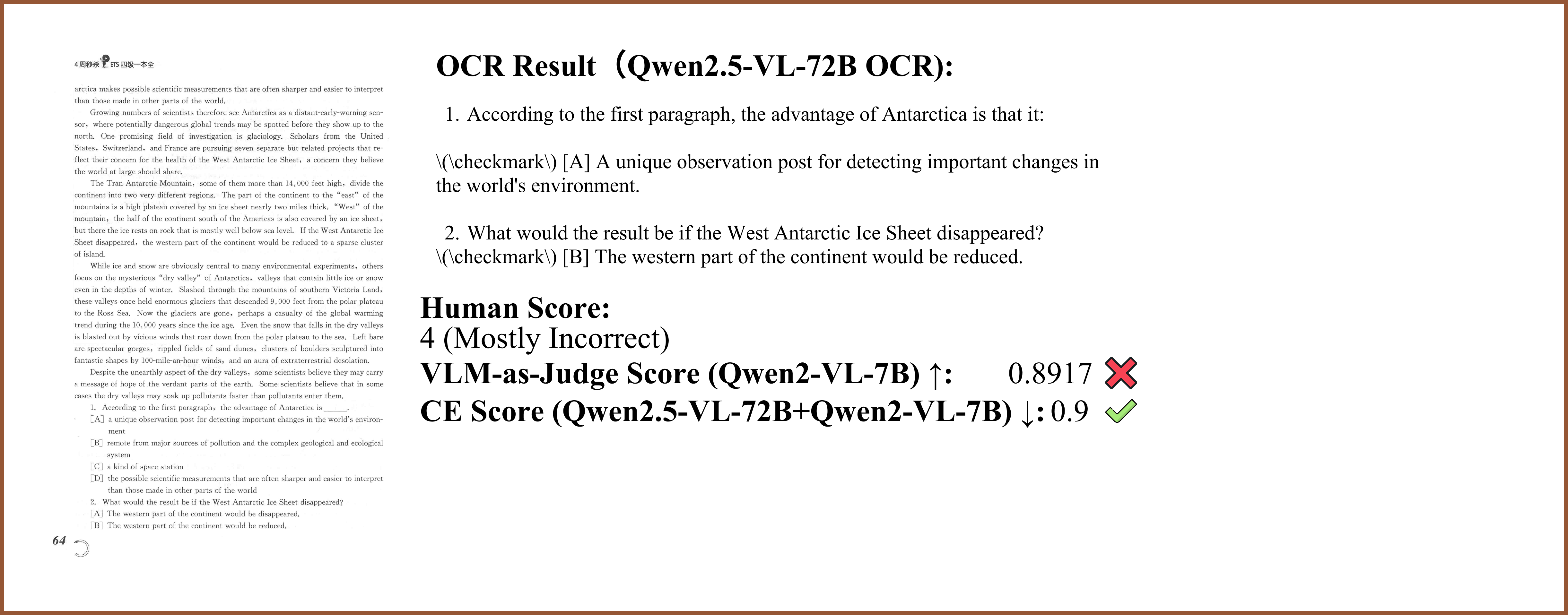}
    \captionof{figure}{Case 5: Multiple-choice test questions accurately evaluated by all methods, showing robust performance on educational content with specialized formatting.}
    \label{fig:case5}
\end{minipage}
\end{tcolorbox}

These cases demonstrate that both CE and VLM-as-Judge methods generally align with human evaluation, but with important differences. CE shows particularly strong correlation with human judgments on complex documents (Cases 1, 3, and 5), where nuanced understanding is required. The strong agreement between human scores and automated methods validates our approach, particularly the training-free CE method which achieves comparable performance to supervised VLM-as-Judge approaches without requiring explicit quality assessment prompting.

Notably, in cases with clear quality issues (Case 2), both methods accurately identify poor OCR results, confirming their reliability for quality filtering applications. For structured content (Case 4), both approaches excel, suggesting that format preservation is well-captured by these evaluation techniques.

These qualitative examples complement our quantitative results in Section 4, reinforcing that CE provides a robust, unsupervised alternative to more complex and computationally expensive VLM-as-Judge methods for OCR quality assessment.

\section{Prompt}
The effectiveness of OCR tasks in VLLMs is significantly influenced by the design of prompts. We present the primary prompts utilized in our experiments below, each serving a specific purpose in the evaluation framework. These prompts were carefully crafted based on extensive preliminary explorations to maximize OCR performance while maintaining consistency across different models.

\begin{tcolorbox}[
    title={OCR VLM-as-Judge Prompt}, 
    breakable, 
    colback=white, 
    colframe=black, 
    fonttitle=\bfseries, 
    fontupper=\footnotesize
]
You are an expert evaluator assessing the quality of OCR (Optical Character Recognition) model predictions. \\
You will receive:

- A question
    
- A prediction generated by the OCR model.
    
- The corresponding image containing text.

Your task is to judge how well the predicted text matches the visual textual content in the image, with respect to the question's intent. \\
\textbf{Evaluation criteria:}
\begin{enumerate}
    \item Focus only on whether the prediction correctly reflects the textual content of the image.
    \item Assign a score from 0 to 1 in steps of 0.1, using the following four-level guideline:
\end{enumerate}

\textbf{Scoring Reference:}

- \textbf{0.9-1.0 (Perfect Match)} \\
    The prediction matches the image text exactly, with no errors or omissions.
    
- \textbf{0.7–0.8 (Minor Errors, Still Clear)} \\
    The prediction is very close to the image text, with only small mistakes that do not affect understanding.
    
- \textbf{0.4–0.6 (Partially Correct)} \\
    The prediction contains noticeable errors or captures only part of the text, reducing clarity.
    
- \textbf{0.0–0.3 (Mostly or Completely Incorrect)} \\
    The prediction is largely incorrect or unrelated to the text in the image.

Respond only with the numerical score (e.g., 0.9). Do not include any explanation or commentary.
\end{tcolorbox}

\begin{tcolorbox}[
    title={OCR CE-Guided Router Prompt}, 
    breakable, 
    colback=white, 
    colframe=black, 
    fonttitle=\bfseries, 
    fontupper=\footnotesize
]
You are an expert AI assistant tasked with improving answers to visual questions. 
Please look at the image and examine the following question and the current answers from different models.

Question: \{question\}

Current model predictions:
\{predictions\_str\}

Your task is to synthesize these predictions and create a single improved answer that:
\begin{enumerate}
    \item Is more accurate based on the visual content
    \item Is concise and direct
    \item Uses a natural, conversational tone
    \item Maintains the core meaning of the original answers if they were correct
    \item Improves clarity and precision
\end{enumerate}

Do not invent details not present in the image. Your answer should be grounded in what is actually visible.

Please provide ONLY the improved answer with no explanations or additional text.
\end{tcolorbox}
\begin{tcolorbox}[
    title={OCR Qwen2.5VL-72B Prompt}, 
    breakable, 
    colback=white, 
    colframe=black, 
    fonttitle=\bfseries, 
    fontupper=\footnotesize
]
You are a very professional expert at OCR tasks. Please analyze the following image and extract all text content. 
\begin{enumerate}
\item Ensure that the extracted text matches the original in the image and maintains the original structure. 
\item If the image contains two columns, you should extract all text in the left column before you move on to the right. 
\item Ignore the headers, but keep the footnotes, title. 
\item You could either use LaTeX, KaTeX or Markdown format for math formulas, physics equations, chemical expressions etc. 
\item Do not add or modify the original content!
\end{enumerate}
\end{tcolorbox}

\subsection{Evaluation Metrics}
To evaluate model performance, we employed multiple metrics assessing various aspects of OCR quality. The character error rate (CER) and word error rate (WER) measured the character-level and word-level accuracy respectively, capturing fine-grained recognition performance. BLEU and ROUGE-L scores assessed the overall textual similarity between predictions and ground truth, with ROUGE-L particularly sensitive to longer n-gram overlaps that indicate structural preservation. For semantic accuracy, we calculated embedding similarity using sentence transformers, which effectively captures meaning preservation even when exact wording differs. Additionally, we employed task-specific metrics for specialized OCR applications: formula recognition used a LaTeX-aware F1 score that accounts for equivalent expressions, while table extraction was evaluated using TEDS (Table Edit Distance Similarity) to measure structural fidelity. Collectively, these metrics provided a comprehensive evaluation framework that assessed both the syntactic accuracy and semantic fidelity of OCR outputs across diverse document understanding scenarios.

\subsection{Details of Compared Methods}
Our comparative analysis incorporated several state-of-the-art approaches for OCR evaluation and improvement. The VLM-as-Judge paradigm was implemented using three different foundation models: GPT4o, Qwen2-VL-72B, and Qwen2-VL-7B, each prompted with the standardized evaluation instructions shown in Section A.2. For traditional evaluation metrics, we utilized the benchmark methodology from OCRBench, employing exact match and fuzzy matching criteria with standardized preprocessing to normalize formatting variations. The self-verification methods included both the token-level confidence based approach and our proposed Consensus Entropy framework. For token-level confidence, we aggregated model-reported logit scores and calibrated them using temperature scaling. The CE framework was implemented with multiple variations in entropy calculation methods, including Mean Distance, Sum, Max, and Mean methods as detailed in Section 3.1 of the main paper. All baseline methods were evaluated using identical test samples and environmental configurations to ensure fair comparison.

\section{Ensemble Results}
\subsection{CCOCR}
Table~\ref{tab:ccocr_comb} summarizes the ensemble performance of three CE‑selected model combinations on the \textsc{CCOCR} benchmark, 
spanning four major OCR task categories: \textit{KIE}, \textit{Document Parsing}, \textit{Multilingual OCR}, and \textit{Multi‑scene OCR}. 
For each combination, we report both the ensemble results (first row) and the performance of the best‑performing single model in the group (second row, \textit{italicized}). 
The final column reports the \textit{Overall Score}, computed as the average across the four tasks.

\begin{table}[h]
  \centering
  \caption{CCOCR Ensemble Results} 
  \begin{tabular}{lccccc}
  \toprule
  \multirow{2}{*}{\makecell{Models}} & \multicolumn{4}{c}{Task Performance} & Overall \\
  \cmidrule(lr){2-5}
  & KIE & Doc Parsing & Multi Language & Multi Scene & Score \\
  \midrule
  Qwen2.5-VL-3B,\ Qwen2.5-VL-7B,\ Qwen2.5-VL-72B & 89.27 & \sethlcolor{tabcolor5}\hl{64.85} & 79.75 & 85.56 & \sethlcolor{tabcolor3}\hl{79.86} \\
  \hspace{1em}\textit{Best Single (Qwen2.5-VL-72B)} & 89.51 & 62.34 & 80.77 & 86.34 & 79.74 \\
  \midrule
  Qwen2.5-VL-3B,\ Qwen2.5-VL-7B,\ Qwen2-VL-7B-Instruct & 87.16 & \sethlcolor{tabcolor5}\hl{61.19} & 77.49 & 83.09 & 77.23 \\
  \hspace{1em}\textit{Best Single (Qwen2.5-VL-7B)} & 87.42 & 60.67 & 78.49 & 84.12 & 77.67 \\
  \midrule
  InternVL2\_5-4B,\ Qwen2.5-VL-7B,\ Qwen2.5-VL-72B & 88.71 & \sethlcolor{tabcolor5}\hl{62.48} & 79.90 & 84.90 & 79.00 \\
  \hspace{1em}\textit{Best Single (Qwen2.5-VL-72B)} & 89.51 & 62.34 & 80.77 & 86.34 & 79.74 \\
  \midrule
  \bottomrule
  \end{tabular}
  \label{tab:ccocr_comb}
  \end{table}

\subsection{OCRBench-V2}
Table~\ref{tab:ocrbenchv2} presents the bilingual evaluation of seven CE‑selected multi‑model aggregation schemes on OCRBench‑V2. 
Each scheme combines four to five vision‑language models (VLMs). 
\textit{Ensemble Score} denotes the overall performance of the aggregated model, 
whereas \textit{Single Best} is the highest‑scoring individual model within the same ensemble for the corresponding language subset. 
Their absolute difference, $\Delta = \textit{Ensemble Score} - \textit{Single Best}$, is reported in the right‑most column. 
Results with the highest score in each language subset are highlighted in blue, 
and all positive gains ($\Delta > 0$) are shaded in green.

\begin{longtable}{lcccccc}
  \caption{OCRBench-V2 Ensemble Results} \\
  \label{tab:ocrbenchv2} \\
  \toprule
  Models & \multicolumn{2}{c}{Ensemble Score} & \multicolumn{2}{c}{Single Best} & \multicolumn{2}{c}{$\Delta$} \\
  \cmidrule(lr){2-3} \cmidrule(lr){4-5} \cmidrule(lr){6-7}
  & English & Chinese & English & Chinese & English & Chinese \\
  \midrule
  \endfirsthead
  
  \multicolumn{7}{c}{continuation sheet \arabic{table}: OCRBench-V2 Ensemble Results} \\
  \toprule
  Models & \multicolumn{2}{c}{Ensemble Score} & \multicolumn{2}{c}{Single Best} & \multicolumn{2}{c}{$\Delta$} \\
  \cmidrule(lr){2-3} \cmidrule(lr){4-5} \cmidrule(lr){6-7}
  & English & Chinese & English & Chinese & English & Chinese \\
  \midrule
  \endhead
  
  \midrule
  \multicolumn{7}{r}{continued on next page} \\
  \endfoot
  
  \bottomrule
  \endlastfoot
  internvl2\ 5\ 26b,\ qwen2vl-8b,\ gemini\ pro\\\hspace{0.5em}MiniCPM-V-2\ 6 & \sethlcolor{tabcolor5}\hl{0.588} & \sethlcolor{tabcolor5}\hl{0.528} & 0.555 & 0.442 & \sethlcolor{tabcolor3}\hl{0.033} & \sethlcolor{tabcolor3}\hl{0.086} \\
  \midrule
  gemini\ pro,\ gpt4o,\ internvl2\ 5\ 26b\\\hspace{0.5em}qwen2vl-8b & \sethlcolor{tabcolor5}\hl{0.589} & \sethlcolor{tabcolor5}\hl{0.521} & 0.555 & 0.442 & \sethlcolor{tabcolor3}\hl{0.034} & \sethlcolor{tabcolor3}\hl{0.079} \\
  \midrule
  internvl2\ 5\ 26b,\ gpt4o,\ qwen2vl-8b\\\hspace{0.5em}MiniCPM-V-2\ 6 & \sethlcolor{tabcolor5}\hl{0.567} & \sethlcolor{tabcolor5}\hl{0.483} & 0.555 & 0.442 & \sethlcolor{tabcolor3}\hl{0.012} & \sethlcolor{tabcolor3}\hl{0.041} \\
  \midrule
  internvl2\_8b,\ cambrian\_8b,\ llava\_onevision\_qwen2\_7b\_ov & \sethlcolor{tabcolor5}\hl{0.538} & \sethlcolor{tabcolor5}\hl{0.395} & 0.404 & 0.363 & \sethlcolor{tabcolor3}\hl{0.134} & \sethlcolor{tabcolor3}\hl{0.032} \\
  \midrule
  idefics3,\ llava\_onevision\_qwen2\_7b\_ov,\ MiniCPM-V-2\_6 & \sethlcolor{tabcolor5}\hl{0.550} & \sethlcolor{tabcolor5}\hl{0.466} & 0.416 & 0.307 & \sethlcolor{tabcolor3}\hl{0.134} & \sethlcolor{tabcolor3}\hl{0.159} \\
  \midrule
  internvl2\_8b,\ cambrian\_8b,\ llavar & \sethlcolor{tabcolor5}\hl{0.521} & \sethlcolor{tabcolor5}\hl{0.387} & 0.404 & 0.363 & \sethlcolor{tabcolor3}\hl{0.117} & \sethlcolor{tabcolor3}\hl{0.024} \\
  \midrule
  internvl2\_8b,\ cambrian\_8b,\ textharmony & \sethlcolor{tabcolor5}\hl{0.520} & \sethlcolor{tabcolor5}\hl{0.394} & 0.404 & 0.363 & \sethlcolor{tabcolor3}\hl{0.116} & \sethlcolor{tabcolor3}\hl{0.031} \\
  \midrule
  \end{longtable}
  
\subsection{OCRBench}
Tables~\ref{OCRBench_com5}, \ref{OCRBench_comb4},\ref{OCRBench_com3}  present supplementary results on \textsc{OCRBench}, covering CE-selected ensemble combinations of 3, 4, and 5 models. Each row represents a specific ensemble configuration. The \textit{Score} column reports the final ensemble performance obtained via CE-based prediction selection. The \textit{Max}, \textit{Min}, and \textit{Avg} columns correspond to the highest, lowest, and mean scores among the participating individual models. The rightmost columns denote the absolute gains of the ensemble over its components: $\Delta_{\text{max}} = \textit{Score} - \textit{Min}$, $\Delta_{\text{min}} = \textit{Score} - \textit{Max}$, and $\Delta_{\text{avg}} = \textit{Score} - \textit{Avg}$. Blue highlights indicate the best result among both the ensemble and its constituent models, while green shading marks positive relative gains.

\begin{longtable}{lcccccccc}
\caption{OCRBench Ensemble Results (5 models)} \\
\label{OCRBench_com5} \\
\toprule
Models & Score & Max & Min & Avg & $\Delta_{max}$ & $\Delta_{min}$ & $\Delta_{avg}$ \\
\midrule
\endfirsthead

\multicolumn{8}{c}{continuation sheet \arabic{table}:OCRBench Ensemble Results (5 models)} \\
\toprule
Models & Score & Max & Min & Avg & $\Delta_{max}$ & $\Delta_{min}$ & $\Delta_{avg}$ \\
\midrule
\endhead

\midrule
\multicolumn{8}{r}{continued on next page} \\
\endfoot

\bottomrule
\endlastfoot
Ovis2-1B, Qwen2.5-VL-72B, SenseChat-Vision\\\hspace{0.5em}Step1V, Step1o & \sethlcolor{tabcolor5}\hl{95.8} & 92.6 & 87.9 & 89.50 & \sethlcolor{tabcolor3}\hl{7.9} & \sethlcolor{tabcolor3}\hl{3.2} & \sethlcolor{tabcolor3}\hl{6.30} \\
\midrule
Ovis2-4B, Qwen2.5-VL-7B, SenseChat-Vision\\\hspace{0.5em}Step1V, Step1o & \sethlcolor{tabcolor5}\hl{95.7} & 92.6 & 87.4 & 89.78 & \sethlcolor{tabcolor3}\hl{8.3} & \sethlcolor{tabcolor3}\hl{3.1} & \sethlcolor{tabcolor3}\hl{5.92} \\
\midrule
Ovis2-4B, Qwen2.5-VL-72B, SenseChat-Vision\\\hspace{0.5em}Step1V, Step1o & \sethlcolor{tabcolor5}\hl{95.6} & 92.6 & 87.9 & 89.88 & \sethlcolor{tabcolor3}\hl{7.7} & \sethlcolor{tabcolor3}\hl{3.0} & \sethlcolor{tabcolor3}\hl{5.72} \\
\midrule
Ovis2-4B, Qwen-VL-Plus-0809, SenseChat-Vision\\\hspace{0.5em}Step1V, Step1o & \sethlcolor{tabcolor5}\hl{95.6} & 92.6 & 84.3 & 89.16 & \sethlcolor{tabcolor3}\hl{11.3} & \sethlcolor{tabcolor3}\hl{3.0} & \sethlcolor{tabcolor3}\hl{6.44} \\
\end{longtable}

\definecolor{tabcolor3}{RGB}{210, 231, 209} 
\definecolor{tabcolor5}{RGB}{212, 220, 248} 

\begin{longtable}{lcccccccc}
\caption{OCRBench Ensemble Results (4 models)} \\
\label{OCRBench_comb4} \\
\toprule
Models & Score & Max & Min & Avg & $\Delta_{max}$ & $\Delta_{min}$ & $\Delta_{avg}$ \\
\midrule
\endfirsthead

\multicolumn{8}{c}{continuation sheet \arabic{table}: OCRBench Ensemble Results (4 models)} \\
\toprule
Models & Score & Max & Min & Avg & $\Delta_{max}$ & $\Delta_{min}$ & $\Delta_{avg}$ \\
\midrule
\endhead

\midrule
\multicolumn{8}{r}{continued on next page} \\
\endfoot

\bottomrule
\endlastfoot
Ovis2-1B, Qwen2.5-VL-7B, Step1V\\\hspace{0.5em}Step1o & \sethlcolor{tabcolor5}\hl{95.5} & 92.6 & 87.4 & 89.40 & \sethlcolor{tabcolor3}\hl{8.1} & \sethlcolor{tabcolor3}\hl{2.9} & \sethlcolor{tabcolor3}\hl{6.10} \\
\midrule
Qwen2.5-VL-72B, SenseChat-Vision, Step1V\\\hspace{0.5em}Step1o & \sethlcolor{tabcolor5}\hl{95.4} & 92.6 & 87.9 & 89.62 & \sethlcolor{tabcolor3}\hl{7.5} & \sethlcolor{tabcolor3}\hl{2.8} & \sethlcolor{tabcolor3}\hl{5.78} \\
\midrule
MUG-U-7B, Qwen2.5-VL-72B, Step1V\\\hspace{0.5em}Step1o & \sethlcolor{tabcolor5}\hl{95.3} & 92.6 & 87.9 & 90.05 & \sethlcolor{tabcolor3}\hl{7.4} & \sethlcolor{tabcolor3}\hl{2.7} & \sethlcolor{tabcolor3}\hl{5.25} \\
\midrule
MUG-U-7B, Qwen-VL-Plus-0809, Step1V\\\hspace{0.5em}Step1o & \sethlcolor{tabcolor5}\hl{95.1} & 92.6 & 84.3 & 89.15 & \sethlcolor{tabcolor3}\hl{10.8} & \sethlcolor{tabcolor3}\hl{2.5} & \sethlcolor{tabcolor3}\hl{5.95} \\
\end{longtable}

\definecolor{tabcolor3}{RGB}{210, 231, 209} 
\definecolor{tabcolor5}{RGB}{212, 220, 248} 

\begin{longtable}{lcccccccc}
\caption{OCRBench Ensemble Results (3 models)} \\
\label{OCRBench_com3} \\
\toprule
Models & Score & Max & Min & Avg & $\Delta_{max}$ & $\Delta_{min}$ & $\Delta_{avg}$ \\
\midrule
\endfirsthead

\multicolumn{8}{c}{continuation sheet \arabic{table}: OCRBench Ensemble Results (3 models)} \\
\toprule
Models & Score & Max & Min & Avg & $\Delta_{max}$ & $\Delta_{min}$ & $\Delta_{avg}$ \\
\midrule
\endhead

\midrule
\multicolumn{8}{r}{continued on next page} \\
\endfoot

\bottomrule
\endlastfoot
Ovis2-8B, SenseChat-Vision, Step1o & \sethlcolor{tabcolor5}\hl{94.1} & 92.6 & 89.3 & 90.43 & \sethlcolor{tabcolor3}\hl{4.8} & \sethlcolor{tabcolor3}\hl{1.5} & \sethlcolor{tabcolor3}\hl{3.67} \\
\midrule
Qwen2.5-VL-7B, Step1V, Step1o & \sethlcolor{tabcolor5}\hl{94.0} & 92.6 & 87.4 & 89.53 & \sethlcolor{tabcolor3}\hl{6.6} & \sethlcolor{tabcolor3}\hl{1.4} & \sethlcolor{tabcolor3}\hl{4.47} \\
\midrule
Ovis2-1B, Step1V, Step1o & \sethlcolor{tabcolor5}\hl{94.0} & 92.6 & 88.6 & 90.07 & \sethlcolor{tabcolor3}\hl{5.4} & \sethlcolor{tabcolor3}\hl{1.4} & \sethlcolor{tabcolor3}\hl{3.93} \\
\midrule
Qwen2-VL-72B-Instruct, Step1V, Step1o & \sethlcolor{tabcolor5}\hl{94.0} & 92.6 & 88.6 & 90.00 & \sethlcolor{tabcolor3}\hl{5.4} & \sethlcolor{tabcolor3}\hl{1.4} & \sethlcolor{tabcolor3}\hl{4.00} \\
\midrule
Ovis2-4B, Qwen2-VL-72B-Instruct, Step1o & \sethlcolor{tabcolor5}\hl{94.0} & 92.6 & 88.8 & 90.77 & \sethlcolor{tabcolor3}\hl{5.2} & \sethlcolor{tabcolor3}\hl{1.4} & \sethlcolor{tabcolor3}\hl{3.23} \\
\midrule
MUG-U-7B, Step1V, Step1o & \sethlcolor{tabcolor5}\hl{93.9} & 92.6 & 88.6 & 90.77 & \sethlcolor{tabcolor3}\hl{5.3} & \sethlcolor{tabcolor3}\hl{1.3} & \sethlcolor{tabcolor3}\hl{3.13} \\
\midrule
MUG-U-7B, Ovis2-4B, Step1o & \sethlcolor{tabcolor5}\hl{93.9} & 92.6 & 90.9 & 91.53 & \sethlcolor{tabcolor3}\hl{3.0} & \sethlcolor{tabcolor3}\hl{1.3} & \sethlcolor{tabcolor3}\hl{2.37} \\
\midrule
MUG-U-7B, Qwen-VL-Plus-0809, Step1o & \sethlcolor{tabcolor5}\hl{93.8} & 92.6 & 84.3 & 89.33 & \sethlcolor{tabcolor3}\hl{9.5} & \sethlcolor{tabcolor3}\hl{1.2} & \sethlcolor{tabcolor3}\hl{4.47} \\
\end{longtable}

\section{Additional Representation Space Analysis}
Additional analysis of VLM output convergence and divergence patterns on 210 models (beyond Figure~\ref{fig:entropy}) will be released with the code and dataset at \url{https://github.com/Aslan-yulong/consensus-entropy}.

\section{Limitations and Future Directions}
Despite the promising results demonstrated by the Consensus Entropy framework, several limitations warrant acknowledgment and suggest avenues for future research. The current implementation relies on semantic embedding spaces that may not fully capture the nuances of specialized domains such as mathematical formulas or chemical notations. Additionally, the framework's performance is contingent on having multiple independent VLMs available at inference time, which may impose computational constraints in resource-limited scenarios. The optimal threshold for routing decisions currently requires empirical calibration for each specific application domain, limiting immediate out-of-the-box deployment.

Future research directions could address these limitations through domain-specific embedding techniques for specialized content types, more efficient ensemble methods requiring fewer models, and adaptive threshold mechanisms that automatically adjust to document characteristics. Additional investigations into the theoretical properties of prediction convergence patterns could lead to more principled frameworks for uncertainty quantification beyond the OCR domain. The insights from these convergence patterns might also inform model design and training objectives, potentially enhancing individual model robustness. Exploring the relationship between model architecture diversity and ensemble effectiveness represents another promising direction, particularly in identifying minimal but complementary model combinations that maximize performance gains while minimizing computational overhead.

\ifdefined\CEAppendixIncluded
\else
\bibliographystyle{ACM-Reference-Format}
\bibliography{main}
\fi


\end{document}